\documentclass[conference]{IEEEtran}

\usepackage{amsmath,amssymb,amsfonts}
\usepackage{graphicx}
\usepackage{siunitx}
\usepackage{natbib}
\setcitestyle{authoryear,open={(},close={)}}
\usepackage{textcomp}
\usepackage{xcolor}
\usepackage{color, colortbl}
\usepackage{multirow}
\usepackage{booktabs}
\renewcommand{\thetable}{\arabic{table}}
\usepackage{algorithm}
\usepackage[noend]{algpseudocode}

\def\BibTeX{{\rm B\kern-.05em{\sc i\kern-.025em b}\kern-.08em
    T\kern-.1667em\lower.7ex\hbox{E}\kern-.125emX}}
\begin{document}

\title{Transfer Learning for Protein Structure Classification at Low Resolution}

\author{\IEEEauthorblockN{Alexander Hudson\textsuperscript{*}}
\IEEEauthorblockA{\textit{School of Electronic Engineering and Computer Science} \\
\textit{Queen Mary University of London}\\
London, UK \\
\textsuperscript{*}\textit{Corresponding author}\\
a.o.hudson@se18.qmul.ac.uk}
\and
\IEEEauthorblockN{Shaogang Gong}
\IEEEauthorblockA{\textit{School of Electronic Engineering and Computer Science} \\
\textit{Queen Mary University of London}\\
London, UK \\
s.gong@qmul.ac.uk}

}

\maketitle

\begin{abstract}
Structure determination is key to understanding protein function at a molecular level. Whilst significant advances have been made in predicting structure and function from amino acid sequence, researchers must still rely on expensive, time-consuming analytical methods to visualise detailed protein conformation. In this study, we demonstrate that it is possible to make accurate ($\geq$80\%) predictions of protein class and architecture from structures determined at low ($>$3\si{\angstrom}) resolution, using a deep convolutional neural network trained on high-resolution ($\leq$3\si{\angstrom}) structures represented as 2D matrices. Thus, we provide proof of concept for high-speed, low-cost protein structure classification at low resolution, and a basis for extension to prediction of function. We investigate the impact of the input representation on classification performance, showing that side-chain information may not be necessary for fine-grained structure predictions. Finally, we confirm that high-resolution, low-resolution and NMR-determined structures inhabit a common feature space, and thus provide a theoretical foundation for boosting with single-image super-resolution.\footnote{Code to accompany this article is available at https://github.com/danobohud/TransferLearningforPSC}\\

\end{abstract}


\section{Introduction}
\begin{figure}[htbp]
\centerline{\includegraphics[width=85mm,height=60mm,keepaspectratio]{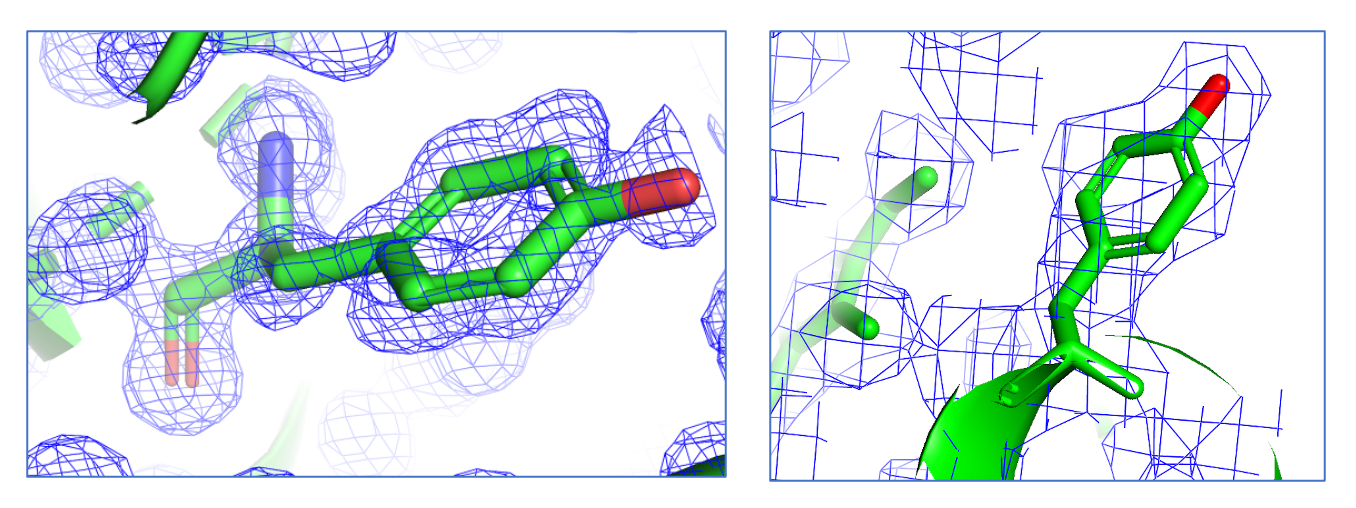}}
\caption{Impact of resolution on structure determination. Myoglobin tyrosine 103 shown from structures 1a6m (1\si{\angstrom}) and 108m (2.7\si{\angstrom}). Electron density displayed as blue cages, atoms shown as sticks. Adapted from PDB 101 \citep{Berman2000} and produced using pyMol \citep{DeLano2020}.}
\label{fig}
\end{figure}

Proteins are large biological molecules consisting of chains of amino acids that are of particular interest to life science research, as they perform a wide variety of essential functions in the cell \citep{Alberts2007}. Functional characterisation of proteins can be arduous, and as such structural biologists can rely on the close relationship between structure and function to predict activity from structure given a known taxonomy of well-characterised protein folds \citep{Whisstock2003}, to complement sequence alignment studies \citep{Eisenhaber2000}. Broadly speaking, the greater the resolution of a solved structure (given in \si{\angstrom}ngstr{\"o}ms, 10$^{-10}$m), the more information can be derived from it: individual atoms can be resolved below 1\si{\angstrom}, the polypeptide backbone and amino acid side-chains under 3\si{\angstrom}, and protein backbone conformation at over 3\si{\angstrom} \citep{Berman2000}, see \textbf{Fig. \ref{fig}}. The need for atomic resolution is reflected in publication bias, with structures determined at $\leq$3\si{\angstrom} currently making up 93\% of the Protein Data Bank (PDB) \citep{Berman2000}. 

Unfortunately, high-resolution structure solving is challenging and represents a fundamental bottleneck in research: to date, more than 120 million amino acid sequences have been determined, but only 160,000 structures have been published  \citep{Berman2000,UniProt}. X-Ray Crystallography (XRC) has historically been the most commonly used technique in protein structure determination, but is time-consuming and expensive: competition for access to facilities is fierce, and costs can reach \$100,000 per structure \citep{Stevens2003}. The requirement for crystallisation also excludes certain protein groups of interest, including some large transmembrane assemblies \citep{Meury2011}. Nuclear Magnetic Resonance (NMR) can yield information not only on topology but on dynamics, but has historically been limited to small soluble proteins \citep{Sugiki2017}. The advent of cryo-Electron Microscopy (cryo-EM) has permitted the visualisation of proteins in near-native conformations at under 2\si{\angstrom}, but the technique remains prohibitively expensive \citep{Peplow2017,Hand2020}. 

As a result, the holy grail of structural biology has become the accurate prediction of protein structure from amino acid sequence alone \citep{Kuhlman2019}. Recent years have seen huge progress in the field, but there remains room for improvement: the best predictors of the most recent Critical Assessment of Structure Prediction competition (CASP13) achieving not more than 80\% accuracy in \textit{ab initio} backbone placement for the most challenging structures \citep{Kryshtafovych2019}.  Furthermore, common metrics of predictive accuracy, such as precision of contact prediction, \textit{GDT\_TS} and \textit{RMSD}, do not take into account amino acid side-chain placement, which is crucial to understanding protein function \citep{Zemla2003,Kryshtafovych2019}. 

\section{Problem statement}

In light of these challenges, before and if sequence-based structure prediction is solved and/or low-cost, high-resolution imaging becomes widely available, the ability to make accurate predictions of protein structure and function from low-resolution data could feasibly accelerate the pace of research. 

Building on previous work \citep{Sikosek2019}, this project sets out to identify whether the features learned by convolutional neural networks (CNNs) trained on 2D representations of high-resolution structures (defined as $\leq$3\si{\angstrom}) can be used accurately to classify fine-grained protein fold topology from structures determined at low resolution ($>$3\si{\angstrom}). Secondly, we seek to identify which form of  input performs best in protein structure classification (PSC), comparing atom selections representative of low, medium and high information content.

\section{Contribution}

We show for the first time that it is feasible to make accurate ($\geq$ 80\%) predictions of protein class and architecture from structures solved at low resolution, including a challenging set determined with NMR. In this way, we provide a theoretical basis for mapping between low- and high-resolution structures, and for extension to function prediction. We find that the best predictors are those trained on matrices encoding distances between C$_{\alpha}$, C$_{\beta}$, oxygen and nitrogen atoms of the protein backbone (\textbf{Fig. \ref{fig1}}), outperforming heavy atom and alpha carbon selections, and so demonstrate the importance of selecting a representation appropriate to the task. Finally, we achieve benchmark classification performance (89\% accuracy)  on  prediction of homologous superfamily from over 5,150 possible categories, using a four-component ensemble of deep CNNs.

\section{Background}
\vspace{2.5mm}
\subsection{Artificial neural networks (ANNs)}\label{back1}
\vspace{2.5mm}

ANNs are a family of machine learning algorithms whose architecture is loosely analogous to the neurons of the mammalian brain, and which have been shown to be powerful predictive tools in disciplines including computer vision and natural language processing \citep{Krizhevsky2012,Devlin2019}. ANNs are composed of sequential layers of simple computational units (\textit{nodes}), in which the output of any node is an elementwise combination of its inputs passed through some  non-linear activation function \citep{Goodfellow2016}. Given sufficient data, the parameters of these models may be learned via back-propagation in response to a training signal \citep{LeCun1988}, enabling ANNs to learn arbitrarily complex predictive functions. The more intermediate or "hidden" layers to a network - \textit{deep} ANNs having two or more such layers -  the more complex the function it can learn, at the cost of greater computational complexity.  

\subsection{Convolution for image classification and transfer learning}\label{back2}
\vspace{2.5mm}
Convolutional neural networks (CNNs) are ANNs containing one or more convolutional layer and which are applied to data with a known grid-like topology, such as images and videos \citep{Goodfellow2016}. The convolution operation allows a layer to scan over its input matrix with a sliding window of stacked nodes (\textit{kernels}), storing the strongest node outputs in an “activation map” via a pooling operation \citep{LeCun1990}. The power of CNNs in image classification has long been recognised: from Yann LeCun’s work on recognising handwritten digits, to the use of deeper networks and innovative model architectures to label images from the ImageNet repository \citep{LeCun1990,Krizhevsky2012,He2016,Simonyan2015}. Subsequent work showed that the discriminatory features learned by these models in one image domain (the \textit{source}) can be transferred to classify data in a separate, noisier or more challenging domain (the \textit{target}). Examples of such \textit{transfer learning} approaches include pre-training a network on an image classification task and fine-tuning on a separate object detection task \citep{Razavian2014}, and simultaneous learning between paired high-  and low-quality images \citep{Chen2015}.

\subsection{Representing proteins as images: protein distance maps}\label{back3}
\vspace{2.5mm}
Computational biologists have profited from these advances by converting publicly available three-dimensional protein structures into two-dimensional \textit{protein distance maps} (hereafter, PDMs): symmetric matrices encoding the pairwise distances between atoms \textit{i} and \textit{j} (a$_{i}$, a$_{j}$) of a solved structure \citep{Phillips1970,Hu2002}.

\begin{figure}[htbp]
\centerline{\includegraphics[width=90mm,height=90mm,keepaspectratio]{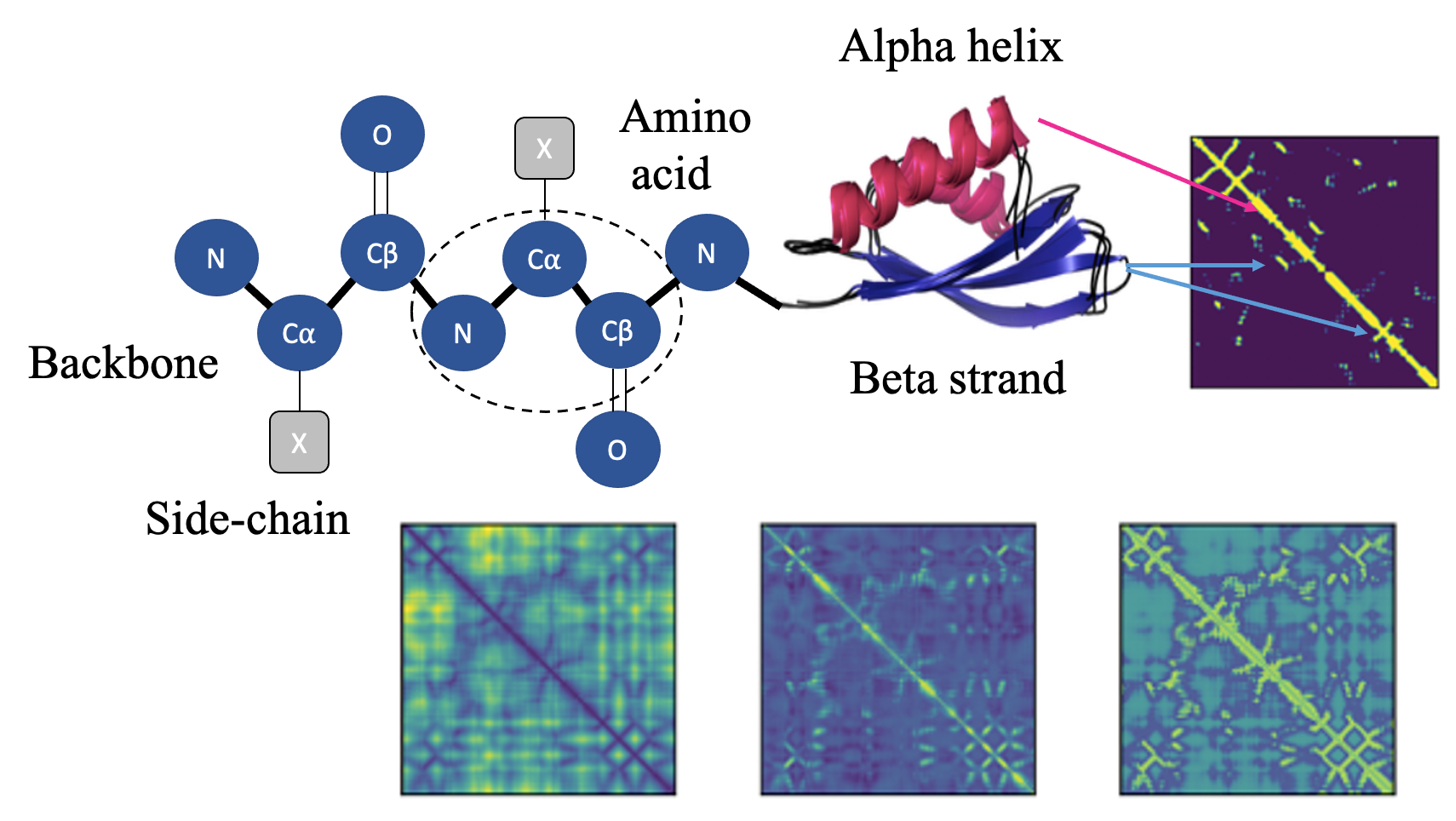}}
\caption{Representing proteins in two dimensions. \textit{Top left}: Maps may be constructed from distances between the alpha carbon (C$_{\alpha}$), beta carbon (C$_{\beta}$), polypeptide backbone (thick black line) or heavy atoms (all non-Hydrogen atoms) of a protein. A single amino acid is shown in the dotted circle. Adapted from \citep{Anand2018}. \textit{Top centre}: Illustrative secondary structure shown from CATH domain 3.30.70.380 \citep{Orengo1997}.  \textit{Top right}: Example contact map. \textit{Bottom row (from left to right)}: Example distance map, anisotropic network model (ANM) and non-bonded (NB) energy matrices \citep{Sikosek2019}.}
\label{fig1}
\end{figure}

PDMs have the advantage over 3D structure representations of reducing both computational load and sensitivity to feature rotation or translation \citep{Sikosek2019}, and are generally presented in one of two common forms (\textbf{Fig. \ref{fig1}}). \textit{Contact maps} are binary matrices wherein two atoms are identified as being in contact if they fall within a set distance of one another, typically 7-8\si{\angstrom} \citep{Duarte2010}. \textit{Distance maps} directly encode the Euclidean distances between atoms of the protein \citep{DeMelo2006,Pietal2015}. The patterns that appear in these maps correspond to characteristic structural elements, for example alpha helices and beta sheets, as pictured in \textbf{Fig. \ref{fig1}}. 

PDMs may be generated from the distances between different selections of atoms, including the alpha (C$_{\alpha}$) and/or beta (C$_{\beta}$) carbons of the polypeptide backbone, or the heavy (non-hydrogen) atoms of the backbone and side-chains. The relative merits of different representations remain disputed: \citet{Duarte2010} concluded that a combination of C$_{\alpha}$ and C$_{\beta}$ atoms outperforms individual components (and particularly C$_{\alpha}$) when reconstructing 3D protein structures from contact maps, whilst C$_{\alpha}$ maps performed better than side-chain geometric centres for enzyme class prediction in a study by  \citet{DaSilveira2009}, and heavy atom representations performed well in a more recent publication from \citet{Newaz2020}. 

Diverse uses have been found for PDMs in computational biology. Key amongst these are protein structure classification (PSC), as in the present study; retrieval of similar proteins \citep{Liu2018}; as an intermediate step in three-dimensional structure prediction from amino acid sequence \citep{Kuhlman2019}; and even in \textit{de novo} protein design \citep{Anand2018}.     

\subsection{Related work: Protein structure classification}\label{back4}
\vspace{2.5mm}

PSC is the task of assigning a candidate structure to one of a set of discrete three-dimensional patterns (\textit{folds}) containing the same arrangement and topology of secondary structural elements \citep{Craven1995}. Common reference taxonomies include the class, fold, superfamily and family hierarchies of the structural classification of proteins (SCOP) dataset \citep{Fox2014}, and the class, architecture, fold and homologous superfamily classifications of CATH \citep{Orengo1997}. Notably, PSC may also serve as a convenient objective for producing vector embeddings of protein structures for use in some secondary task \citep{Sikosek2019}, the implications of which are explored in \textbf{Section \ref{Limitations}}.

\definecolor{Gray}{gray}{0.9}

A review of the literature was conducted to identify historic approaches to PSC, detailed in \textbf{Appendix Table A\ref{taba1}}. Three broad methodologies were encountered: those in which traditional machine learning algorithms were applied to features extracted from PDMs \citep{Shi2009,Taewijit2010,Vani2016,Pires2011}; a second set training deep CNNs directly on large datasets of maps \citep{Sikosek2019,Eguchi2020}; and ensemble models combining different approaches \citep{Zacharaki2017,Newaz2020}. Studies relying on features derived from amino acid sequence alone are not listed exhaustively, however state of the art is included in \textbf{Table A\ref{taba1}} for completeness \citep{Xia2017,Hou2018}.

Many early PSC studies extracted features from subsets of non-redundant structures labelled according to SCOP class and fold, mining secondary structural features from distance maps using hand-crafted algorithms \citep{Shi2009,Vani2016}.  The best-performing of these \citep{Pires2011} extracted frequency statistics of C$_{\alpha}$ distances and applied K-Nearest Neighbour classifier or Random Forest classifiers to these features, achieving 94\% on prediction of SCOP family. 

\begin{figure}[htbp]
\centerline{\includegraphics[width=85mm,height=60mm,keepaspectratio]{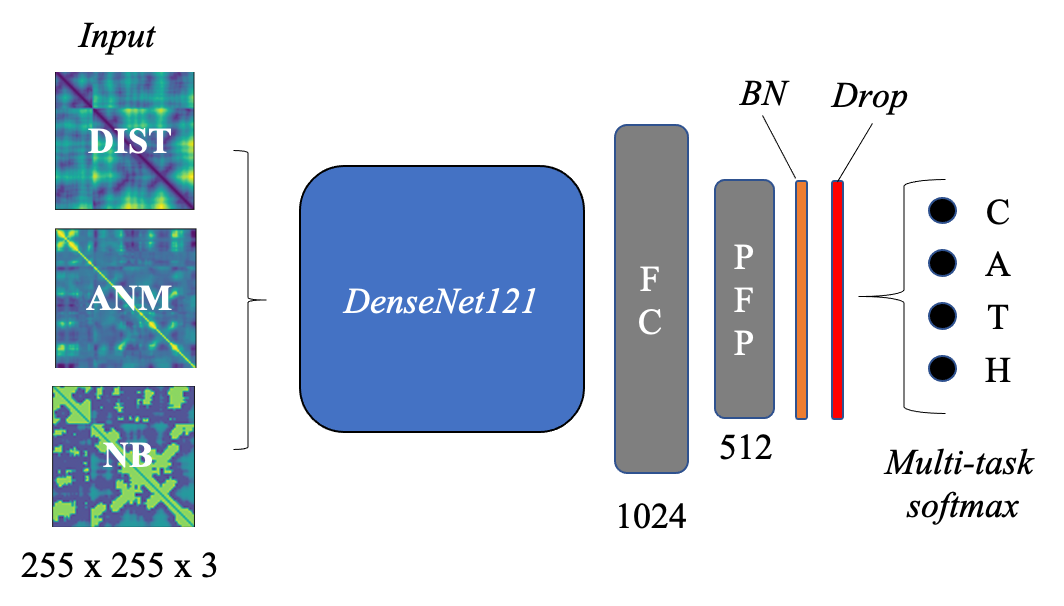}}
\caption{Architecture of the model. \underline{Abbreviations}. DIST: Distance matrix; ANM: Anisotropic network model; NB: Non-bonded energy; FC: Fully connected (Dense) layer; PFP: protein fingerprint; BN: batch normalisation; Drop: Dropout. Channel depth is shown below the input, FC and PFP layers.}
\label{fig2}
\end{figure}

Among the best results in CATH classification have been those achieved using deep CNNs \citep{Sikosek2019,Eguchi2020} and ensembles \citep{Newaz2020}. A modified version of \textit{DenseNet121}, capable of simultaneous multi-class, multi-label prediction of CATH categories, demonstrated up to 87\% accuracy on the most challenging task, being prediction of homologous superfamily from over 2000 possible classes \citep{Sikosek2019}. This model was trained on heavy atom distance maps augmented with measures of intrinsic molecular motion and non-bonded energy, as described in \citep{Sikosek2019}, illustrated in \textbf{Fig. \ref{fig1}} (\textit{bottom row}) and detailed below.  The resultant model was subsequently used to produce \textit{protein fingerprints}, efficient feature vectors produced by the penultimate layer of the trained CNN (\textbf{Fig. \ref{fig2}}) and used in a subsequent step as the input to a random forest prediction of a secondary task: small molecule binding activity as measured by ChEMBL.

\citet{Eguchi2020} deployed a six-layer CNN  with up-sampling and deconvolution for semantic segmentation (pixelwise labelling) of C$_{\alpha}$ distance maps. Applying their model to a CATH non-redundant dataset augmented with cropping and sub-sampled to balance class representation, this group achieved up to 88\% per structure accuracy of architecture prediction. It is important to note that the primary aim of the study was not accurate structure-level classification, but rather labelling individual amino acids according to CATH architecture, achieving an impressive average accuracy of 91\%. 

\citet{Newaz2020} combined models trained on different representations into ensembles for PSC. Among these, distances between the heavy atoms in a protein structure were described as ordered sub-graphs (\textit{graphlets}), whose frequencies then served as an input feature for logistic regression. This study reported 93\%-100\% per-class accuracy on CATH homologous superfamily prediction when combining graphlet, sequence and Tuned Gaussian Interval (GIT) representations. It is important to note that only those classes and sub-classes with thirty or more instances were included in the analysis. This permitted a statistically meaningful comparison of different feature inputs and methods; however, the resultant accuracies may not be representative of performance across the universe of possible folds.  

\section{Methods}
\vspace{2.5mm}

\subsection{Datasets}\label{datasets}
\vspace{2.5mm}

Protein domains from the CATH non-redundant domain set (v4.2) \citep{Orengo1997} were assigned to one of three groups according to resolution: “High-resolution” (HR, $\leq$3\si{\angstrom}), “Low-Resolution” (LR, $>$3\si{\angstrom}) and “NMR” (characterised using solution or solid-state NMR). Unlike in previous studies, instances were not excluded on the basis of chain length or a minimum per class frequency, in order to expose the models to as many representative folds as possible. This variance is reflected in the small differences in frequency of T and H classes between datasets (\textbf{Table A\ref{taba8}}). The impact of class imbalance on model performance is discussed in \textbf{\ref{atomclass}} and \textbf{\ref{Limitations}}).

\subsection{Pre-processing}\label{preprocessing}
Structure files from each group were parsed to obtain a stack of distance map, anisotropic network model (ANM) and non-bonded  energy (NB) matrices, following \citep{Sikosek2019} and described in \textbf{Algorithm \ref{alg1}}. In this way, each image passed to the network encoded information relating to spatial positioning, flexibility and non-bonded energy potentials between each atom of the matrix, respectively. 

Initially, PDB files were processed with ABS-PDB2PQR \citep{Jurrus2018} using an AMBER forcefield, as detailed in \citet{Sikosek2019}. Euclidean distance, ANM cross-correlation and NB matrices were then extracted from PQR files for each domain with ProDy \citep{Bakan2011}, using either alpha carbon (CA),  backbone (BB) or  heavy atom  (HEAVY) selections  (see \textbf{Fig. \ref{fig1}} and \textbf{Algorithm 1}). These atom selections were taken to be representative of low, medium and high information content, respectively: CA maps included the distances between C$_{\alpha}$ atoms only; BB selections including information from C$_{\alpha}$, C$_{\beta}$, oxygen and nitrogen; and heavy atom selections including distances between all non-hydrogen atoms, inclusive of side-chains. Summary statistics for each of the nine datasets are provided in  \textbf{Table A\ref{taba8}}. 

Distance matrices were clipped at a maximum distance of 50\si{\angstrom} (three standard deviations from the mean distance across all maps), and ANM matrices between -1 and +1, before rescaling distance, ANM and NB matrices (in the range [0, 100], [-100, 100] and [-1000, 1000] respectively) for memory-efficient storage. All representations were reshaped with bicubic interpolation and stacked to give a set of 255x255x3-dimensional matrices, comparable to the three channels of RGB colour images.

\begin{algorithm}[h]
  \scriptsize\small
  \caption{High- to Low-Resolution Domain Transfer}\label{alg1}
  \begin{algorithmic}
  \State
    \State \textit{\% Extract matrices from structure files \%}
    \Procedure{Parse}{domainlist, atomgroup}
    \For{domainID \textbf{in} domainlist}
        \State struct $\gets$ \textit{parsePQR}(\textit{pdb2Pqr}(domainID))
        \State atoms $\gets$ \textit{selectAtoms}(struct, atomgroup)
        \State ANM $\gets$ \textit{ANM}(\textit{crossCorrelate}(atoms))
        \State charges $\gets$ \textit{getCharge}(atoms)
        \For{i \textbf{in range} \textit{length}(atoms)}
            \For{j \textbf{in range} \textit{length}(atoms)}
                \State 
                dist[i,j] $\gets$ \textit{Euclidean}(atoms[i],
                atoms[j])
                \State
                NB[i,j] $\gets$ \textit{getNB}(charges[i,j],
                atoms[i,j])
            \EndFor
        \EndFor
    
        \State dist $\gets$ \textit{Resize}(\textit{Clip}(dist,0,50),(255,255)*100)
        \State ANM $\gets$ \textit{Resize}(\textit{Clip}(ANM,-1,1),(255,255)*100)
        \State NB $\gets$ \textit{Resize}(\textit{NB},(255,255)*1000)
        \State \Return{[dist,ANM,NB], domainID}
    \EndFor
    \State
    \EndProcedure
    \textit{\% Train CNN ensemble and return best prediction\%}
    \Procedure{Ensemble}{HRInputs, Labels, TestSet}
    \For{d, Dataset \textbf{in} HRInputs}
        \State TrainedModel $\gets$ \textit{Train}(CNN, Dataset, Labels)
        \State Ensemble[d] $\gets$ TrainedModel
    \EndFor
    \State w $\gets$ 1$/$ \textit{length}(Ensemble)
    \For{model \textbf{in} Ensemble}
        \State P$_{test}$[model] $\gets$ w$*$\textit{Predict}(model, TestSet)
    \EndFor
    \State Predicted $\gets$ $\underset{c\in{C},p\in{P}}{\arg\max}$ P$_{test}$\\
    \State \Return{Predicted}
    \State 
    \EndProcedure
    \State
    \textit{\% High- to low-resolution domain transfer \%}
    \Procedure{Transfer}{HRdomains, TestDomains}
    \For{atom in CA, BB, HEAVY}
    \State HRInputs, Labels $\gets$ PARSE(HRDomains, atom) 
    \EndFor
    \State TestSet,TestLabels $\gets$PARSE(TestDomains,atomgroup)
    \State Predicted $\gets$ ENSEMBLE(HRInputs, Labels, TestSet)
    \State Accuracy, F1 $\gets$ \textit{Evaluate}(Predicted,TestLabels)
    \EndProcedure
  \end{algorithmic}
  
\end{algorithm}

\subsection{Model Architecture}
\vspace{2.5mm}

\textbf{Fig. \ref{fig2}} describes the architecture of the deep CNN used in this study, a modified version of pre-trained \textit{DenseNet121} from Keras \citep{Chollet2015}, adapted from \citep{Sikosek2019}. The final layer of the off-the shelf Keras model was replaced with a single fully-connected protein fingerprint (PFP) layer of 512 dimensions, followed by batch normalisation and dropout layers for regularisation of learned features. The output of these layers was then passed to four parallel softmax activation layers corresponding to the 4 (Class; Task C), 41 (Architecture; Task A), 1391 (Topology; Task T) and 6070 (Homologous Superfamily; Task H) possible categories of the CATH dataset. This framework was adopted with deployment in mind; however, it should be noted that a maximum of 1276/1391 T and 5150/6070 H classes were included in the training set (\textbf{Table A\ref{taba8}}). A simple 5-layer CNN was also constructed for comparison, the details of which are included in  \textbf{Fig. A\ref{figa1}}.

\subsection{Training}\label{training}

Model training and optimisation studies were performed on one of the three high-resolution datasets (HRCA, HRBB, HRHEAVY), with the aim of maximising test time performance on the most challenging classification task (H). 10\% of each high-resolution dataset was retained as a test set for evaluation. 

All models were trained to minimise categorical cross entropy loss for a maximum of 150 epochs on a single NVIDIA Tesla T4 GPU, using a 40\% validation ratio, shuffled batches of 32 instances and 25\% dropout. The initial learning rate was set at 0.001 using an Adam optimiser with no early stopping, and learning rate reduction of 20\% enabled after a plateau of 5 epochs, to a minimum of 0.0001. 

\subsection{Evaluation}\label{back5}
\vspace{2.5mm}

In order to determine the impact of atom selection on performance, predictive accuracy of models trained on HRCA, HRBB or HRHEAVY datasets was first assessed on held-out test data from the corresponding high-resolution test set, such that a model trained on a high-resolution backbone (HRBB) training set would be evaluated on the HRBB test set. The performance of the best of these models, DenseNet121 trained on HRBB (\textit{DN\_HRBB}) was then evaluated on the high-resolution test sets from other atom selections (HRCA, HRHEAVY) and on the entire low-resolution and NMR datasets (LRCA, LRBB, LRHEAVY, NMRCA, NMRBB and NMRHEAVY). 

In addition to accuracy, best model performance was assessed using the F1-score, a harmonic mean of precision and recall that takes account of per-class performance, and the PFP homogeneity score proposed by \citet{Sikosek2019}. For the latter, the quality of feature vectors extracted from the PFP layer of trained models (see \textbf{Fig. \ref{fig2}}) was evaluated by clustering instances with \textit{K-means} \citep{MacQueen1967} according to \textit{k} possible classes for a given task, and comparing the overlap of actual and best predicted label clusters using the homogeneity score functionality of \textit{scikit-learn} \citep{Pedregosa2011}. Best clusters were identified after 10 iterations following initialisation with \textit{kmeans++}.

Finally, the best models from each atom selection (\textit{DN\_HRCA}, \textit{DN\_HRBB} and \textit{DN\_HRHEAVY}) were combined into an ensemble (\textit{DN\_E1}), giving each component an equally weighted vote and assigning the most confident weighted prediction as the predicted label, evaluating on all nine test sets. A second four-member ensemble (\textit{DN\_E2}) was also developed that incorporated an additional model trained on distance only HRBB inputs (\textit{DN\_HRBB\_DIST}).

\section{Results}

Average test time accuracy for models trained and tested on HRCA, HRBB and HRHEAVY data is presented in \textbf{Fig. \ref{fig3}} and \textbf{Table A\ref{taba2}}. To assess the contribution of ANM and NB layers to model performance, a model was trained on a modified HRBB dataset comprising triplicate stacks of distance matrices, shown in \textbf{Fig. \ref{fig3}} (\textit{HRBB\_DIST}) and \textbf{Table A\ref{taba2}}. Performance of the best (\textit{DN\_HRBB}) model on HR (held-out), LR and NMR test sets for all three atom selections is presented in (\textbf{Fig. \ref{HRBBbest}} and \textbf{Table A\ref{taba3}}). The results of combining the best models into weighted ensembles is presented in (\textbf{Tables \ref{tab1}}, \textbf{A\ref{taba4}} and \textbf{A\ref{taba5}}).

\subsection{Impact of atom selection on classification performance}\label{atomclass}
\vspace{2.5mm}

\begin{figure}[htbp]
\centerline{\includegraphics[width=85mm,height=80mm,keepaspectratio]{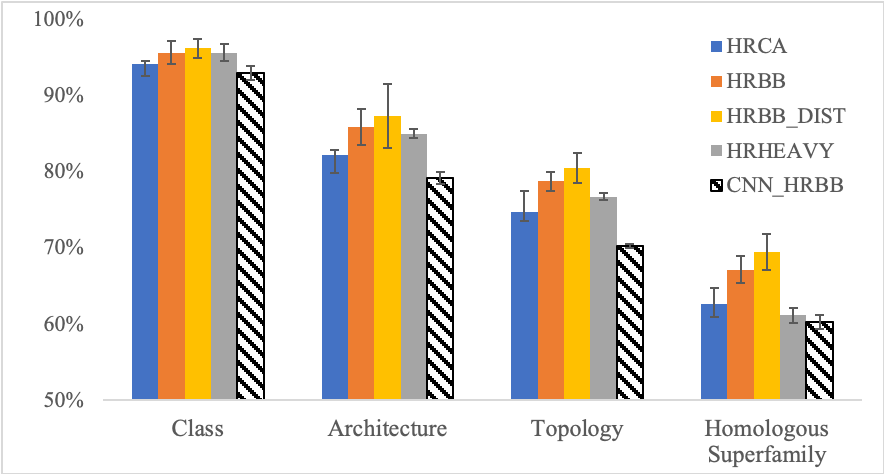}}
\caption{Mean accuracy of models trained and tested on HRCA, HRBB, or HRHEAVY datasets. Error bars show 95\% confidence interval. CA: Alpha carbon; BB: Backbone; BB\_DIST: triplicate stack of distance matrices; HEAVY: heavy atom selection; CNN\_HRBB: CNN comparator, see \textbf{Fig A\ref{figa1}}.}
\label{fig3}
\end{figure}

\textbf{Fig. \ref{fig3}} and \textbf{Table A\ref{taba2}} confirm the finding of \citet{Sikosek2019} that model performance overall correlates with complexity of the task for all atom selections, with highest accuracy seen for task C and worst for task H on all test sets. Across all four tasks, models trained on HRBB maps outperformed those trained on HRCA and HRHEAVY atom selections. For task H, inspection of 95\% confidence intervals showed this difference to be statistically significant in both cases (mean accuracy of 67\% for HRBB, p$<$0.05, N=3). Average  accuracy was numerically lower for HRHEAVY (61\%) than HRCA (63\%) atom selections, but not significantly so.  F1 scores (\textbf{Table A\ref{taba2}}) followed a similar trend, but were generally 1-4\% lower than the corresponding accuracies, indicating an adverse impact on model performance of class imbalance (detailed in \textbf{Table A\ref{taba7}}).  

\begin{table*}[ht]
\begin{center}
\caption{Best performance for \textit{DN\_HRBB}, \textit{DN\_E1} and \textit{DN\_E2}}

\begin{tabular}{|c|c c c c|c c c c|c c c c|}
\hline

\multirow{2}{*}{\textbf{Dataset}} & \multicolumn{4}{c|}{\textbf{\textit{DN\_HRBB}}} & \multicolumn{4}{c|}{\textbf{\textit{DN\_E1}}} & \multicolumn{4}{c|}{\textbf{\textit{DN\_E2}}} \\ 
& C & A & T & H & C & A & T & H & C & A & T & H \\
 \hline

HRCA & 98\% & 92\% & 89\% & 84\% & 96\% & 92\% & 90\% & 84\% & 96\% & 92\% & 90\% & 84\% \\ 
HRBB & 96\% & 86\% & 79\% & 81\% & 94\% & 90\% & 86\% & 80\% & \underline{\textbf{96\%}} & \underline{\textbf{93\%}} & \underline{\textbf{92\%}} & \underline{\textbf{89\%}} \\ 
HRHEAVY & 56\% & 26\% & 16\% & 1\% & 94\% & 76\% & 63\% & 58\% & 94\% & 76\% & 63\% & 58\% \\ 
\hline
LRCA & 93\% & 80\% & 67\% & 51\% & \underline{\textbf{90\%}} & \underline{\textbf{80\%}} & \underline{\textbf{69\%}} & \underline{\textbf{53\%}} & 81\% & 54\% & 44\% & 39\% \\ 
LRBB & 92\% & 79\% & 64\% & 48\% & 87\% & 78\% & 66\% & 49\% & 79\% & 52\% & 42\% & 37\% \\ 
LRHEAVY & 38\% & 13\% & 9\% & 1\% & 89\% & 64\% & 50\% & 43\% & 29\% & 3\% & 40\% & 40\% \\ 
\hline
NMRCA & 91\% & 79\% & 63\% & 46\% & \underline{\textbf{88\%}} &\underline{\textbf{80\%}}  & \underline{\textbf{65\%}} & \underline{\textbf{47\%}} & 78\% & 56\% & 39\% & 33\% \\ 
NMRBB & 91\% & 79\% & 61\% & 44\% & 86\% & 79\% & 62\% & 44\% & 83\% & 57\% & 41\% & 34\% \\ 
NMRHEAVY & 38\% & 7\% & 3\% & 1\% & 88\% & 64\% & 46\% & 36\% & 28\% & 6\% & 38\% & 32\% \\ 
\hline
\multicolumn{5}{|c|}{\textbf{Comparators}} & \multicolumn{8}{c|}{\cellcolor{gray}}\\
\cline{1-5}
Benchmark \citep{Sikosek2019} & 99\% & 95\% & 92\% & 87\% & \multicolumn{8}{c|}{\cellcolor{gray}} \\ 
CNN\_HRBB & 93\% & 79\% & 70\% & 60\% & \multicolumn{8}{c|}{\cellcolor{gray}} \\
\hline
\multicolumn{13}{|c|}{\textit{DN\_HRBB}: DenseNet121 trained on HRBB; \textit{DN\_E1}: Ensemble 1; \textit{DN\_E2}: Ensemble 2; CNN\_HRBB: CNN comparator,}\\
\multicolumn{13}{|c|}{ see \textbf{Fig A\ref{figa1}}. Best performers are underlined for HR (\textit{DN\_E2}), LR (\textit{DN\_E1}) and NMR (\textit{DN\_E1}) test sets.}\\

\hline
\end{tabular}\label{tab1}
\end{center}
\end{table*}

Replacing the ANM and NB layers of HRBB instances with copies of the distance matrix layer improved average accuracy marginally across tasks when compared with a distance-ANM-NB stack (\textbf{Fig. \ref{fig3}} and \textbf{Table A\ref{taba2}}). However, this improvement (69\% vs. 67\% for task H) was not found to be statistically significant. 

\begin{figure}[h]
\centerline{\includegraphics[width=80mm,height=70mm,keepaspectratio]{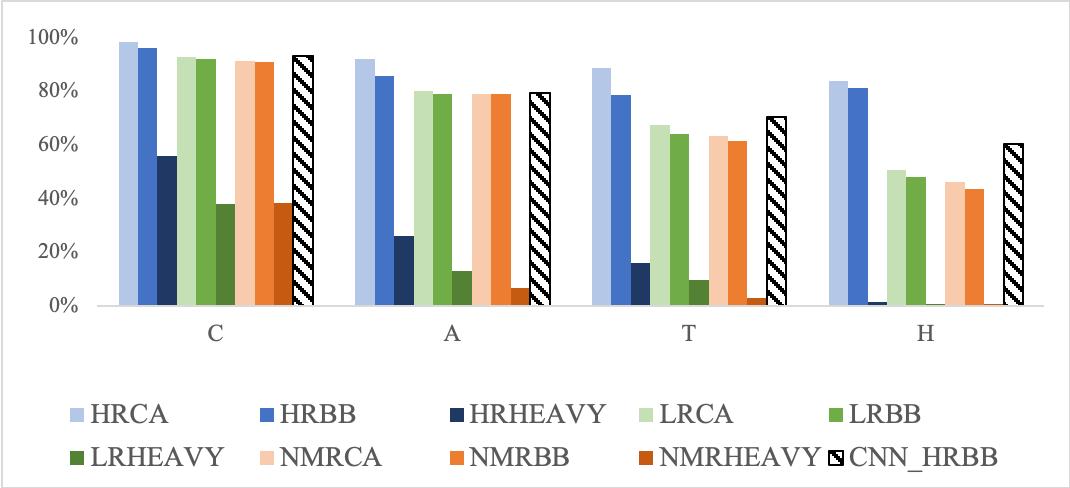}}
\caption{Accuracy of \textit{DN\_HRBB} on C, A, T and H tasks across test sets. CNN\_HRBB: CNN comparator.}
\label{HRBBbest}
\end{figure}

When evaluating trained models on task H with held-out high-resolution data (\textbf{Fig. \ref{HRBBbest}}, \textbf{Tables \ref{tab1}} and \textbf{A\ref{taba3}}), the best (\textit{DN\_HRBB}) model performed better on the HRCA dataset (84\% for task H) than on HRBB data (81\%), and was unable to make predictions from HRHEAVY data (1\%). The former figures compare well with 87\% accuracy in the benchmark study \citep{Sikosek2019}, despite having markedly more classes to choose from (5,150 vs 2,714, \textbf{Table A\ref{taba1}}). A 5-layer CNN, trained and tested on HRBB data (\textit{CNN\_HRBB}, \textbf{Fig. A\ref{figa1}}), performed worse than the average for corresponding \textit{DN\_HRBB} model across all tasks.

PFP homogeneity correlated broadly with accuracy and F1 scores for the high-resolution test sets: learned embeddings produced by \textit{DN\_HRBB} form clusters close to their true labels when provided with inputs from HRCA (84\% homogeneity for task C) and HRBB (83\%) datasets, but not for HRHEAVY (0\%) (\textbf{Table A\ref{taba3}}). PFP homogeneity was consistently lower for the present study when compared with benchmark experiments \citep{Sikosek2019}, but followed a similar trend across tasks.

Application of a random forest classifier to the 512-dimensional protein fingerprints produced by \textit{DN\_HRBB} from each test set did not improve classification performance compared to DenseNet121 alone (\textbf{Table A\ref{taba6}}). 

\subsection{Performance of trained models on low-resolution and NMR datasets}\label{lowresnmr}
\vspace{2.5mm}

The best-performing single model (\textit{DN\_HRBB}) is able to make predictions of over 91\%, 79\%, 63\% and 46\% for class, architecture, topology and homologous superfamily across LRCA and NMRCA datasets (\textbf{Fig. \ref{HRBBbest}} and \textbf{Table A\ref{taba3}}). Predictions were consistently better for LR than for NMR datasets across all test sets. As for HR test sets, performance of \textit{DN\_HRBB} was better for CA than for BB test sets, and was very poor (1\%) for heavy atom selections. Accuracy and F1 scores corresponded very closely for these analyses, and PFP homogeneity scores followed a similar trend to those observed for tests on HR datasets (\textbf{Table A\ref{taba3}}). Any attempts to fine-tune trained models for improved performance on low-resolution or NMR datasets led to loss of performance.

\subsection{Ensemble models}\label{ensembles}
\vspace{2.5mm}

A mixed ensemble of HRCA, HRBB and HRHEAVY models (\textit{DN\_E1}) is able to make task H predictions from HR, LR and NMR data  that are similar to or better than the best HRBB-only model across all classes and atom selections (\textbf{Tables \ref{tab1}} and \textbf{ A\ref{taba4}}). Inclusion of both mixed (distance, ANM and NB) and distance-only representations (model \textit{DN\_E2}, \textbf{Tables \ref{tab1}} and \textbf{ A\ref{taba5}}) improved performance on the HR datasets - up to 89\% accuracy on HRBB (87\% F1) - but damaged predictions on the LR and NMR test sets when compared with E1. 

\section{Discussion}

\subsection{Models trained on backbone selections outperform those trained on alpha carbon or heavy atoms}\label{discuss1}
\vspace{2.5mm}

\textit{DN\_HRBB} achieved up to 84\% accuracy in homologous superfamily prediction on high-resolution test sets (\textbf{Tables \ref{tab1}} and \textbf{A\ref{taba3}}). This compares well not only with benchmark CATH prediction accuracy from distance maps (87\%, over fewer classes), but with sequence-dependent prediction algorithms such as DeepSF, which achieved 75\% test accuracy on the 1175 folds of SCOP1.75 \citep{Hou2018} (\textbf{Table A\ref{taba1}}). 

Models trained on HRBB data performed better than HRCA or HRHEAVY  equivalents (\textbf{Fig. \ref{fig3}} and \textbf{Table A\ref{taba2}}) when testing on held-out data from the same high-resolution dataset. This is unsurprising when comparing backbone representations with the more compact C$_{\alpha}$ representations, and falls in line with the findings of \citet{Duarte2010}. One might expect models trained on heavy atom selections to perform better, as they contain additional information on the relative spatial orientation of side-chain atoms in addition to the carbon, oxygen and nitrogen atoms of the polypeptide backbone (\textbf{Fig. \ref{fig1}}). However, this information is not necessarily required for CATH classification, which is defined using secondary structural characterisation (topology of the backbone), combined with functional annotation using SwissProt \citep{Orengo1997}. As an additional benefit, HRBB representations occupy on average 20\% of the memory of HRHEAVY equivalents before pre-processing (\textbf{Table A\ref{taba8}}). 

The comparative reduction in performance between HRBB and HRHEAVY-trained models could possibly be attributed to loss of representative images during parsing from PDB source structures (HRHEAVY contains 1,932 fewer training instances). An alternative explanation is information loss during rescaling, the average matrix containing 1,234 atoms for heavy and 635 atoms for BB datasets (\textbf{Table A\ref{taba8}}). Reshaping matrices to 255x255 therefore imposes a 23- and 6-fold reduction in area, respectively, compared with a 3-fold upscale for CA instances. 

\subsection{Complex representations may not be required for accurate fold classification}\label{discuss2}

Ablation experiments that removed the ANM and NB layers of the input representation did not significantly impact the point accuracy of predictions made when compared with a distance-ANM-NB stack, but did seem to increase the variance of both accuracy and F1 (\textbf{Table A\ref{taba2}}). This implies that the distance matrix plays a dominant role in model training, but that a more varied input may result in improvements to the diversity (and so robustness) of learned features, as illustrated in the domains triggering maximal activation in the first layer (\textbf{Fig. A\ref{fig4}}).  

\begin{figure*}[htbp]
\centerline{\includegraphics[width=139mm,height=139mm,keepaspectratio]{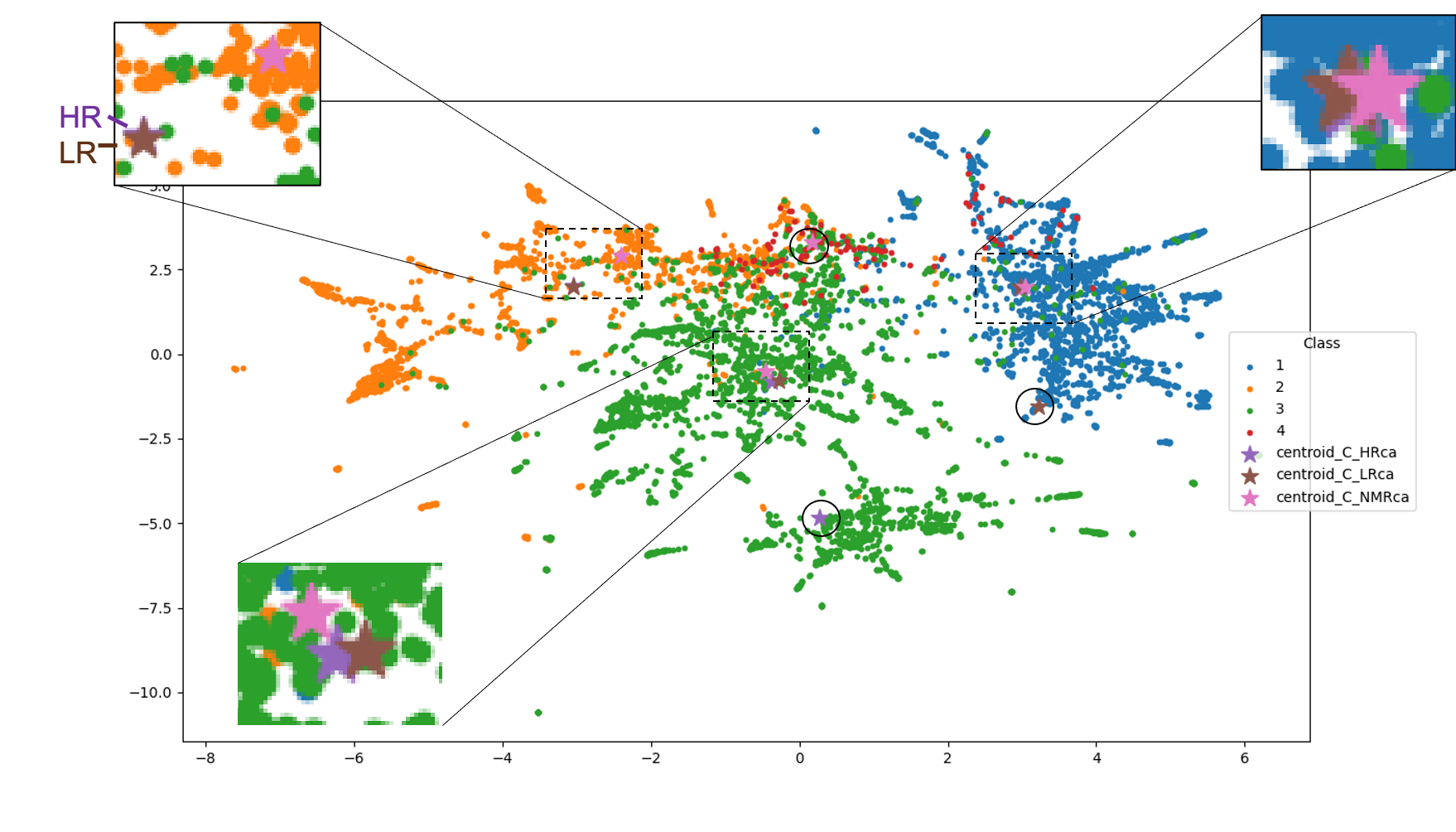}}
\caption{Co-localisation of cluster centroids for HR, LR and NMR datasets. Coloured dots: instances consolidated from HR, LR and NMR datasets and coloured according to class. Stars: cluster centroids. Black circles: misplaced centroids for class 4.}
\label{fig5}
\end{figure*}

\textit{DN\_HRBB} is able to make more accurate predictions from HRCA than from HRBB data (84\% vs. 81\%, \textbf{Fig. \ref{HRBBbest}}, \textbf{Tables \ref{tab1}} and \textbf{A\ref{taba3}}). This suggests a shared feature space between the two datasets, presumably the relative position of the alpha carbons in the {C$_{\alpha}$, C$_{\beta}$, O, N} repeating unit of the polypeptide backbone. This signifies that one can train the classifier using  (BB) representations of intermediate complexity, and deploy on compact (CA) representations whilst simultaneously improving performance. The same is not true of heavy atom selections, where performance of HRBB drops to 1\% (\textbf{Table \ref{tab1}}), possibly as the inclusion of side-chain distances masks the distinctive signals between C$_{\alpha}$ and C$_{\beta}$ atoms. Whilst side-chain information might not be required for accurate CATH classification, it may be useful where learned embeddings are transferred to some secondary task such as prediction of functional site location \citep{Buturovic2014}, small molecule binding \citep{Sikosek2019} or structure retrieval \citep{Liu2018}. Heavy atom selections should therefore not be discounted until the relationship between the input representation and the training objective is fully characterised.

\subsection{Models trained on HR data can be used to make  fold predictions from LR and NMR data}\label{discuss4}
\vspace{2.5mm}

As expected from results with high-resolution data, \textit{DN\_HRBB} performance is better for C, A and T than for H tasks for both low-resolution and NMR datasets. For C and A tasks in particular, accuracies are only marginally worse than seen for the HR test sets (\textbf{Fig \ref{HRBBbest}}) and are improved further using ensembles (\textbf{Table \ref{tab1}}). Whilst one might expect reasonably accurate predictions from structures determined at 3-4\si{\angstrom} as in the LR datasets (\textbf{Table A\ref{taba8}}), the performance of the classifier on NMR structures is more surprising where, as a non-diffraction method, resolution is commonly low ($>$4\si{\angstrom}) or unspecified (999\si{\angstrom}). Further, the nature of the atomic coordinates differs, being an average over an ensemble of possible structures for NMR, and a point estimate based on electron density for XRC and cryo-EM stuctures \citep{Berman2000}.

The ability of the trained model accurately to predict protein class and architecture from LR and NMR test sets is likely attributable to shared patterns of interatomic distances between datasets. To test this hypothesis, 512-dimensional protein fingerprints produced by \textit{DN\_HRBB} were compared for HRCA, LRCA and NMRCA test sets, by computing cluster centroids (class-specific averages) using \textit{K-means}, and transforming the resultant vectors into two dimensions with t-SNE \citep{VanDerMaaten2008}, shown in \textbf{Fig. \ref{fig5}}. 

Comparing the distribution of transformed embeddings (dots, all datasets combined) and centroids (stars, cluster averages for individual datasets) shows that HRCA, LRCA and NMRCA centroids co-localise for classes 1-3 (\textit{mainly $\alpha$}, \textit{mainly $\beta$}, and \textit{$\alpha$-$\beta$}) but not for class 4 (\textit{few secondary structures}). For the dominant classes (1-3), HR (purple) and LR (brown) centroids are generally closer together, and NMR centroids (pink) are close to but generally separate from HR/LR equivalents. The latter may reflect the different methodology for structure determination by NMR, or the composition of proteins suitable for this technique, being generally small (an average of 94 residues, \textbf{Table A\ref{taba8}}) and soluble \citep{Berg2002}. Class 4 instances (shown in red) are underrepresented and exhibit significant overlap with classes 1-3: The K-means algorithm therefore fails to identify them as a discrete cluster (centroids in black circles). This is perhaps unsurprising as the minor class is made up of irregular domains with little secondary structure \citep{Orengo1997}. Class imbalance and overlap for class 4 is reflected in weighted average F1 scores (52\% vs. 94\%-97\%, \textbf{Table A\ref{taba7}}).

\subsection{A multi-model ensemble achieves benchmark performance on high-resolution datasets}\label{discuss5}
\vspace{2.5mm}

A weighted ensemble (\textit{DN\_E2}) of models is able not only to outperform single model-equivalents (\textbf{Table \ref{tab1}}), but achieves 89\% accuracy (87\% F1) on class H prediction from HRBB data, a marginal improvement on the benchmark \citep{Sikosek2019}. Comparing the results achieved for \textit{DN\_E1} and \textit{DN\_E2} illustrates that ensemble components can be tuned to perform on different tasks and input representations, \textit{DN\_E1} performing better on low-resolution data  and E2 on high-resolution data (\textbf{Table \ref{tab1}}).

\section{Limitations and further work}\label{Limitations}

The present study has shown that sequence and side-chain information may not be required for accurate prediction of structure classification at high-resolution. However, it should be noted that PSC often serves as a convenient training objective to produce embeddings as an input for some secondary task, particularly in identifying related domains \citep{Liu2018}, rather than as the primary task \textit{per se}. A crucial extension of this work is therefore to assess the impact of including side-chain information and/or sequence information on performance in secondary tasks, for example on the "TAPE" tasks developed by \citet{Rao2019}. 

Class imbalance is a well-known challenge in protein structure classification \citep{Vani2016}. In previous studies, datasets have been carefully trimmed in order to balance representation, for example by including only those classes with thirty or more representative structures \citep{Newaz2020}. This approach was discounted in the present study, training on all available instances from the CATH non-redundant dataset in order to maximise coverage of the universe of possible classes. As a result, many categories of superfamily are represented by a single domain, and the vast majority (92-98\%) of superfamilies contain fewer than ten instances (\textbf{Table A\ref{taba8}}). This imbalance generates a risk that trained models not be able correctly to classify new unseen minority class instances, which could be assessed in future studies by testing model performance using other datasets such as SCOP \citep{Fox2014}. Possible techniques to counteract class imbalance include boosting minority representation in the training set with additional structures drawn from PDB, synthetic minority oversampling (SMOTE) as in \citep{Vani2016}, or sub-cropping \citep{Eguchi2020}. Other possible avenues to explore include objective function re-weighting \citep{Eguchi2020}, weighted ensembles of class-specific models, and minority class incremental rectification \citep{Dong2019}. 

We have shown that it is possible to make accurate ($\geq$80\%) predictions of protein class (C) and architecture (A) from low-resolution and even NMR data, but that performance drops significantly for the more challenging topology and homologous superfamily tasks. One possible approach to improving performance on low-resolution structures is to integrate representations of the same class obtained using different experimental methods into individual instances, an example of multi-view learning \citep{Zhao2017}. 

Finally, the evidence presented confirms that low-resolution and NMR structures inhabit a common feature space with high-resolution data, and so provides a theoretical basis for mapping between the domains using techniques such as single image super-resolution \citep{Dong2016}. Such a mapping could help to overcome the bottleneck in obtaining high-resolution structures and so accelerate the pace of future research into human health and disease. 

\section*{Acknowledgments}
{The authors would like to thank the following: Dr. Tobias Sikosek (formerly of GSK) for his advice on representations and model design, and for review of this manuscript; Professor Possu Huang and Dr Namrata Anand (Stanford University) for sharing datasets; Professor Adrian Shepherd (Birkbeck University of London) for his advice on potential applications; and Professor David Harris (Oxford University) for proofreading this article.}

\bibliography{Bibliography}

\begin{thebibliography}{57}
\providecommand{\natexlab}[1]{#1}
\providecommand{\url}[1]{\texttt{#1}}
\expandafter\ifx\csname urlstyle\endcsname\relax
  \providecommand{\doi}[1]{doi: #1}\else
  \providecommand{\doi}{doi: \begingroup \urlstyle{rm}\Url}\fi

\bibitem[Alberts et~al.(2007)Alberts, Johnson, Lewis, Raff, Roberts, and
  Walter]{Alberts2007}
Bruce Alberts, Alexander Johnson, Julian Lewis, Martin Raff, Keith Roberts, and
  Peter Walter.
\newblock \emph{{Molecular Biology of the Cell}}.
\newblock Garland Science, New York, 4th edition, 2007.
\newblock \doi{10.1201/9780203833445}.

\bibitem[Anand and Huang(2018)]{Anand2018}
Namrata Anand and Possu Huang.
\newblock {Workshop track-ICLR 2018 Generative Modeling for Protein
  Structures}.
\newblock \emph{Iclr}, 2018.

\bibitem[Bakan et~al.(2011)Bakan, Meireles, and Bahar]{Bakan2011}
Ahmet Bakan, Lidio~M. Meireles, and Ivet Bahar.
\newblock {ProDy: Protein dynamics inferred from theory and experiments}.
\newblock \emph{Bioinformatics}, 2011.
\newblock ISSN 13674803.
\newblock \doi{10.1093/bioinformatics/btr168}.

\bibitem[{Berg JM, Tymoczko JL}(2002)]{Berg2002}
Stryer~L. {Berg JM, Tymoczko JL}.
\newblock {Three-Dimensional Protein Structure Can Be Determined by NMR
  Spectroscopy and X-Ray Crystallography}.
\newblock \emph{Biochemistry}, 2002.

\bibitem[Berman et~al.(2000)Berman, Westbrook, Feng, Gilliland, Bhat, Weissig,
  Shindyalov, and Bourne]{Berman2000}
Helen~M. Berman, John Westbrook, Zukang Feng, Gary Gilliland, T.~N. Bhat, Helge
  Weissig, Ilya~N. Shindyalov, and Philip~E. Bourne.
\newblock {The Protein Data Bank}, 2000.
\newblock ISSN 03051048.

\bibitem[Buturovic et~al.(2014)Buturovic, Wong, Tang, Altman, and
  Petkovic]{Buturovic2014}
Ljubomir Buturovic, Mike Wong, Grace~W. Tang, Russ~B. Altman, and Dragutin
  Petkovic.
\newblock {High precision prediction of functional sites in protein
  structures}.
\newblock \emph{PLoS ONE}, 2014.
\newblock ISSN 19326203.
\newblock \doi{10.1371/journal.pone.0091240}.

\bibitem[Chen et~al.(2015)Chen, Huang, Feris, Brown, Dong, and Yan]{Chen2015}
Qiang Chen, Junshi Huang, Rogerio Feris, Lisa~M. Brown, Jian Dong, and
  Shuicheng Yan.
\newblock {Deep domain adaptation for describing people based on fine-grained
  clothing attributes}.
\newblock In \emph{Proceedings of the IEEE Computer Society Conference on
  Computer Vision and Pattern Recognition}, 2015.
\newblock ISBN 9781467369640.
\newblock \doi{10.1109/CVPR.2015.7299169}.

\bibitem[Chollet(2015)]{Chollet2015}
Fran{\c{c}}ois Chollet.
\newblock {Keras: The Python Deep Learning library}.
\newblock \emph{Keras.Io}, 2015.

\bibitem[Craven et~al.(1995)Craven, Mural, Hauser, and Uberbacher]{Craven1995}
M.~W. Craven, R.~J. Mural, L.~J. Hauser, and E.~C. Uberbacher.
\newblock {Predicting protein folding classes without overly relying on
  homology.}
\newblock \emph{Proceedings / ... International Conference on Intelligent
  Systems for Molecular Biology ; ISMB. International Conference on Intelligent
  Systems for Molecular Biology}, 1995.
\newblock ISSN 15530833.

\bibitem[{Da Silveira} et~al.(2009){Da Silveira}, Pires, Minardi, Ribeiro,
  Veloso, Lopes, Meira, Neshich, Ramos, Habesch, and Santoro]{DaSilveira2009}
Carlos~H. {Da Silveira}, Douglas~E.V. Pires, Raquel~C. Minardi, Cristina
  Ribeiro, Caio~J.M. Veloso, Julio~C.D. Lopes, Wagner Meira, Goran Neshich,
  Carlos~H.I. Ramos, Raul Habesch, and Marcelo~M. Santoro.
\newblock {Protein cutoff scanning: A comparative analysis of cutoff dependent
  and cutoff free methods for prospecting contacts in proteins}.
\newblock \emph{Proteins: Structure, Function and Bioinformatics}, 2009.
\newblock ISSN 08873585.
\newblock \doi{10.1002/prot.22187}.

\bibitem[{De Melo} et~al.(2006){De Melo}, Lopes, Fernandes, {Da Silveira},
  Santoro, Carceroni, Meira, and Ara{\'{u}}jo]{DeMelo2006}
Raquel~C. {De Melo}, Carlos Eduardo~R. Lopes, Fernando~A. Fernandes,
  Carlos~Henrique {Da Silveira}, Marcelo~M. Santoro, Rodrigo~L. Carceroni,
  Wagner Meira, and Arnaldo De~A. Ara{\'{u}}jo.
\newblock {A contact map matching approach to protein structure similarity
  analysis}.
\newblock \emph{Genetics and Molecular Research}, 2006.
\newblock ISSN 16765680.

\bibitem[DeLano(2020)]{DeLano2020}
W~L DeLano.
\newblock {The PyMOL Molecular Graphics System, Version 2.3}, 2020.
\newblock ISSN 1348-4214.

\bibitem[Devlin et~al.(2019)Devlin, Chang, Lee, and Toutanova]{Devlin2019}
Jacob Devlin, Ming~Wei Chang, Kenton Lee, and Kristina Toutanova.
\newblock {BERT: Pre-training of deep bidirectional transformers for language
  understanding}.
\newblock In \emph{NAACL HLT 2019 - 2019 Conference of the North American
  Chapter of the Association for Computational Linguistics: Human Language
  Technologies - Proceedings of the Conference}, 2019.
\newblock ISBN 9781950737130.

\bibitem[Dong et~al.(2016)Dong, Loy, He, and Tang]{Dong2016}
Chao Dong, Chen~Change Loy, Kaiming He, and Xiaoou Tang.
\newblock {Image Super-Resolution Using Deep Convolutional Networks}.
\newblock \emph{IEEE Transactions on Pattern Analysis and Machine
  Intelligence}, 2016.
\newblock ISSN 01628828.
\newblock \doi{10.1109/TPAMI.2015.2439281}.

\bibitem[Dong et~al.(2019)Dong, Gong, and Zhu]{Dong2019}
Qi~Dong, Shaogang Gong, and Xiatian Zhu.
\newblock {Imbalanced Deep Learning by Minority Class Incremental
  Rectification}.
\newblock \emph{IEEE Transactions on Pattern Analysis and Machine
  Intelligence}, 2019.
\newblock ISSN 19393539.
\newblock \doi{10.1109/TPAMI.2018.2832629}.

\bibitem[Duarte et~al.(2010)Duarte, Sathyapriya, Stehr, Filippis, and
  Lappe]{Duarte2010}
Jose~M. Duarte, Rajagopal Sathyapriya, Henning Stehr, Ioannis Filippis, and
  Michael Lappe.
\newblock {Optimal contact definition for reconstruction of Contact Maps}.
\newblock \emph{BMC Bioinformatics}, 2010.
\newblock ISSN 14712105.
\newblock \doi{10.1186/1471-2105-11-283}.

\bibitem[Eguchi and Huang(2020)]{Eguchi2020}
Raphael~R. Eguchi and Po~Ssu Huang.
\newblock {Multi-scale structural analysis of proteins by deep semantic
  segmentation}.
\newblock \emph{Bioinformatics (Oxford, England)}, 2020.
\newblock ISSN 13674811.
\newblock \doi{10.1093/bioinformatics/btz650}.

\bibitem[Eisenhaber(2000)]{Eisenhaber2000}
Frank Eisenhaber.
\newblock {Prediction of Protein Function Two Basic Concepts and One Practical
  Recipe}.
\newblock \emph{Database [Internet]. Austin (TX Landes Bioscience}, 2000.

\bibitem[Fox et~al.(2014)Fox, Brenner, and Chandonia]{Fox2014}
Naomi~K. Fox, Steven~E. Brenner, and John~Marc Chandonia.
\newblock {SCOPe: Structural Classification of Proteins - Extended, integrating
  SCOP and ASTRAL data and classification of new structures}.
\newblock \emph{Nucleic Acids Research}, 2014.
\newblock ISSN 03051048.
\newblock \doi{10.1093/nar/gkt1240}.

\bibitem[Goodfellow et~al.(2016)Goodfellow, Bengio, Courville, and
  Courville]{Goodfellow2016}
Ian Goodfellow, Yoshua Bengio, Aaron Courville, and Aaron Courville.
\newblock \emph{{Deep Learning}}.
\newblock The MIT Press, London, England, 2016.
\newblock ISBN 978-0262035613.
\newblock \doi{10.1038/nmeth.3707}.
\newblock URL \url{www.deeplearningbook.org}.

\bibitem[Hand(2020)]{Hand2020}
Eric Hand.
\newblock {‘We need a people's cryo-EM.' Scientists hope to bring
  revolutionary microscope to the masses}.
\newblock \emph{Science}, 2020.
\newblock ISSN 0036-8075.
\newblock \doi{10.1126/science.aba9954}.

\bibitem[He et~al.(2016)He, Zhang, Ren, and Sun]{He2016}
Kaiming He, Xiangyu Zhang, Shaoqing Ren, and Jian Sun.
\newblock {Deep residual learning for image recognition}.
\newblock In \emph{Proceedings of the IEEE Computer Society Conference on
  Computer Vision and Pattern Recognition}, 2016.
\newblock ISBN 9781467388504.
\newblock \doi{10.1109/CVPR.2016.90}.

\bibitem[Hou et~al.(2018)Hou, Adhikari, and Cheng]{Hou2018}
Jie Hou, Badri Adhikari, and Jianlin Cheng.
\newblock {DeepSF: Deep convolutional neural network for mapping protein
  sequences to folds}.
\newblock \emph{Bioinformatics}, 2018.
\newblock ISSN 14602059.
\newblock \doi{10.1093/bioinformatics/btx780}.

\bibitem[Hu et~al.(2002)Hu, Shen, Shao, Bystroff, and Zaki]{Hu2002}
Jingjing Hu, Xiaolan Shen, Yu~Shao, Chris Bystroff, and Mohammed~J Zaki.
\newblock {Mining Protein Contact Maps}.
\newblock \emph{Proceedings of the 2nd International Conference on Data Mining
  in Bioinformatics}, 2002.

\bibitem[Jurrus et~al.(2018)Jurrus, Engel, Star, Monson, Brandi, Felberg,
  Brookes, Wilson, Chen, Liles, Chun, Li, Gohara, Dolinsky, Konecny, Koes,
  Nielsen, Head-Gordon, Geng, Krasny, Wei, Holst, McCammon, and
  Baker]{Jurrus2018}
Elizabeth Jurrus, Dave Engel, Keith Star, Kyle Monson, Juan Brandi, Lisa~E.
  Felberg, David~H. Brookes, Leighton Wilson, Jiahui Chen, Karina Liles, Minju
  Chun, Peter Li, David~W. Gohara, Todd Dolinsky, Robert Konecny, David~R.
  Koes, Jens~Erik Nielsen, Teresa Head-Gordon, Weihua Geng, Robert Krasny,
  Guo~Wei Wei, Michael~J. Holst, J.~Andrew McCammon, and Nathan~A. Baker.
\newblock {Improvements to the APBS biomolecular solvation software suite}.
\newblock \emph{Protein Science}, 2018.
\newblock ISSN 1469896X.
\newblock \doi{10.1002/pro.3280}.

\bibitem[Krizhevsky et~al.(2012)Krizhevsky, Sutskever, and
  Hinton]{Krizhevsky2012}
Alex Krizhevsky, Ilya Sutskever, and Geoffrey~E. Hinton.
\newblock {ImageNet classification with deep convolutional neural networks}.
\newblock In \emph{Advances in Neural Information Processing Systems}, 2012.
\newblock ISBN 9781627480031.

\bibitem[Kryshtafovych et~al.(2019)Kryshtafovych, Schwede, Topf, Fidelis, and
  Moult]{Kryshtafovych2019}
Andriy Kryshtafovych, Torsten Schwede, Maya Topf, Krzysztof Fidelis, and John
  Moult.
\newblock {Critical assessment of methods of protein structure prediction
  (CASP)—Round XIII}, 2019.
\newblock ISSN 10970134.

\bibitem[Kuhlman and Bradley(2019)]{Kuhlman2019}
Brian Kuhlman and Philip Bradley.
\newblock {Advances in protein structure prediction and design}, 2019.
\newblock ISSN 14710080.

\bibitem[le~Cun(1988)]{LeCun1988}
Yann le~Cun.
\newblock {A theoretical framework for Back-Propagation}.
\newblock In \emph{Proceedings of the 1988 Connectionist Models Summer School},
  1988.

\bibitem[LeCun et~al.(1990)LeCun, Boser, Denker, Henderson, Howard, Hubbard,
  and Jackel]{LeCun1990}
Yann LeCun, Bernhard~E. Boser, John~S. Denker, Donnie Henderson, R.~E. Howard,
  Wayne~E. Hubbard, and Lawrence~D. Jackel.
\newblock Handwritten digit recognition with a back-propagation network.
\newblock In D.~S. Touretzky, editor, \emph{Advances in Neural Information
  Processing Systems 2}, pages 396--404. Morgan-Kaufmann, 1990.

\bibitem[Liu et~al.(2018)Liu, Ye, Wang, and Peng]{Liu2018}
Yang Liu, Qing Ye, Liwei Wang, and Jian Peng.
\newblock {Learning structural motif representations for efficient protein
  structure search}.
\newblock In \emph{Bioinformatics}, volume 34(17), pages i773--i780. Oxford
  University Press, sep 2018.
\newblock \doi{10.1093/bioinformatics/bty585}.

\bibitem[MacQueen(1967)]{MacQueen1967}
J~B MacQueen.
\newblock {Kmeans and Analysis of Multivariate Observations}.
\newblock \emph{5th Berkeley Symposium on Mathematical Statistics and
  Probability 1967}, 1967.
\newblock ISSN 00970433.
\newblock \doi{citeulike-article-id:6083430}.

\bibitem[Meury et~al.(2011)Meury, Harder, Ucurum, Boggavarapu, Jeckelmann, and
  Fotiadis]{Meury2011}
Marcel Meury, Daniel Harder, Z{\"{o}}hre Ucurum, Rajendra Boggavarapu,
  Jean~Marc Jeckelmann, and Dimitrios Fotiadis.
\newblock {Structure determination of channel and transport proteins by
  high-resolution microscopy techniques}.
\newblock \emph{Biological Chemistry}, 2011.
\newblock ISSN 14316730.
\newblock \doi{10.1515/BC.2011.004}.

\bibitem[Newaz et~al.(2020)Newaz, Ghalehnovi, Rahnama, Antsaklis, and
  Milenkovi{\'{c}}]{Newaz2020}
Khalique Newaz, Mahboobeh Ghalehnovi, Arash Rahnama, Panos~J. Antsaklis, and
  Tijana Milenkovi{\'{c}}.
\newblock {Network-based protein structural classification}.
\newblock \emph{Royal Society Open Science}, 2020.
\newblock ISSN 2054-5703.
\newblock \doi{10.1098/rsos.191461}.

\bibitem[Nguyen et~al.(2014)Nguyen, Shang, and Xu]{Nguyen2014}
Son~P. Nguyen, Yi~Shang, and Dong Xu.
\newblock {DL-PRO: A novel deep learning method for protein model quality
  assessment}.
\newblock In \emph{Proceedings of the International Joint Conference on Neural
  Networks}, 2014.
\newblock ISBN 9781479914845.
\newblock \doi{10.1109/IJCNN.2014.6889891}.

\bibitem[Orengo et~al.(1997)Orengo, Michie, Jones, Jones, Swindells, and
  Thornton]{Orengo1997}
C.~A. Orengo, A.~D. Michie, S.~Jones, D.~T. Jones, M.~B. Swindells, and J.~M.
  Thornton.
\newblock {CATH - A hierarchic classification of protein domain structures}.
\newblock \emph{Structure}, 1997.
\newblock ISSN 09692126.
\newblock \doi{10.1016/s0969-2126(97)00260-8}.

\bibitem[Pedregosa et~al.(2011)Pedregosa, Varoquaux, Gramfort, Michel, Thirion,
  Grisel, Blondel, Prettenhofer, Weiss, Dubourg, Vanderplas, Passos,
  Cournapeau, Brucher, Perrot, and Duchesnay]{Pedregosa2011}
Fabian Pedregosa, Gael Varoquaux, Alexandre Gramfort, Vincent Michel, Bertrand
  Thirion, Olivier Grisel, Mathieu Blondel, Peter Prettenhofer, Ron Weiss,
  Vincent Dubourg, Jake Vanderplas, Alexandre Passos, David Cournapeau,
  Matthieu Brucher, Matthieu Perrot, and {\'{E}}douard Duchesnay.
\newblock {Scikit-learn: Machine learning in Python}.
\newblock \emph{Journal of Machine Learning Research}, 2011.
\newblock ISSN 15324435.

\bibitem[Peplow(2017)]{Peplow2017}
Mark Peplow.
\newblock {Cryo-electron microscopy makes waves in pharma labs}.
\newblock \emph{Nature Reviews Drug Discovery}, 2017.
\newblock ISSN 14741784.
\newblock \doi{10.1038/nrd.2017.240}.

\bibitem[Phillips(1970)]{Phillips1970}
D.~C. Phillips.
\newblock {The development of crystallographic enzymology.}, 1970.
\newblock ISSN 00678694.

\bibitem[Pietal et~al.(2015)Pietal, Bujnicki, and Kozlowski]{Pietal2015}
Michal~J. Pietal, Janusz~M. Bujnicki, and Lukasz~P. Kozlowski.
\newblock {GDFuzz3D: A method for protein 3D structure reconstruction from
  contact maps, based on a non-Euclidean distance function}.
\newblock \emph{Bioinformatics}, 2015.
\newblock ISSN 14602059.
\newblock \doi{10.1093/bioinformatics/btv390}.

\bibitem[Pires et~al.(2011)Pires, de~Melo-Minardi, dos Santos, da~Silveira,
  Santoro, and Meira]{Pires2011}
Douglas~E.V. Pires, Raquel~C. de~Melo-Minardi, Marcos~A. dos Santos, Carlos~H.
  da~Silveira, Marcelo~M. Santoro, and Wagner Meira.
\newblock {Cutoff Scanning Matrix (CSM): Structural classification and function
  prediction by protein inter-residue distance patterns}.
\newblock \emph{BMC Genomics}, 2011.
\newblock ISSN 14712164.
\newblock \doi{10.1186/1471-2164-12-S4-S12}.

\bibitem[Rao et~al.(2019)Rao, Bhattacharya, Thomas, Duan, Chen, Canny, Abbeel,
  and Song]{Rao2019}
Roshan Rao, Nicholas Bhattacharya, Neil Thomas, Yan Duan, Xi~Chen, John Canny,
  Pieter Abbeel, and Yun~S. Song.
\newblock {Evaluating Protein Transfer Learning with TAPE}.
\newblock \emph{ArXiv}, jun 2019.
\newblock URL \url{http://arxiv.org/abs/1906.08230}.

\bibitem[Razavian et~al.(2014)Razavian, Azizpour, Sullivan, and
  Carlsson]{Razavian2014}
Ali~Sharif Razavian, Hossein Azizpour, Josephine Sullivan, and Stefan Carlsson.
\newblock {CNN features off-the-shelf: An astounding baseline for recognition}.
\newblock In \emph{IEEE Computer Society Conference on Computer Vision and
  Pattern Recognition Workshops}, 2014.
\newblock ISBN 9781479943098.
\newblock \doi{10.1109/CVPRW.2014.131}.

\bibitem[Shi and Zhang(2009)]{Shi2009}
Jian~Yu Shi and Yan~Ning Zhang.
\newblock {Fast SCOP classification of structural class and fold using
  secondary structure mining in distance matrix}.
\newblock In \emph{Lecture Notes in Computer Science (including subseries
  Lecture Notes in Artificial Intelligence and Lecture Notes in
  Bioinformatics)}, 2009.
\newblock ISBN 3642040306.
\newblock \doi{10.1007\/978-3-642-04031-3\_30}.

\bibitem[Sikosek(2019)]{Sikosek2019}
Tobias Sikosek.
\newblock {Protein structure featurization via standard image classification
  neural networks}.
\newblock \emph{bioRxiv}, page 841783, nov 2019.
\newblock \doi{10.1101/841783}.
\newblock URL \url{https://doi.org/10.1101/841783}.

\bibitem[Simonyan and Zisserman(2015)]{Simonyan2015}
Karen Simonyan and Andrew Zisserman.
\newblock {Very deep convolutional networks for large-scale image recognition}.
\newblock In \emph{3rd International Conference on Learning Representations,
  ICLR 2015 - Conference Track Proceedings}, 2015.

\bibitem[Stevens(2003)]{Stevens2003}
R.~C Stevens.
\newblock {The cost and value of three-dimensional protein structure}.
\newblock \emph{Drug Discovery World}, 4\penalty0 (3):\penalty0 35--48, 2003.

\bibitem[Sugiki et~al.(2017)Sugiki, Kobayashi, and Fujiwara]{Sugiki2017}
Toshihiko Sugiki, Naohiro Kobayashi, and Toshimichi Fujiwara.
\newblock {Modern Technologies of Solution Nuclear Magnetic Resonance
  Spectroscopy for Three-dimensional Structure Determination of Proteins Open
  Avenues for Life Scientists}, 2017.
\newblock ISSN 20010370.

\bibitem[Taewijit and Waiyamai(2010)]{Taewijit2010}
Siriwon Taewijit and Kitsana Waiyamai.
\newblock {CM-HMM: Inter-residue contact and HMM-profiles based enzyme
  subfamily prediction and structure analysis}.
\newblock In \emph{Proceedings of the 9th IEEE International Conference on
  Cognitive Informatics, ICCI 2010}, 2010.
\newblock ISBN 9781424480401.
\newblock \doi{10.1109/COGINF.2010.5599792}.

\bibitem[UniProt(2015)]{UniProt}
UniProt.
\newblock {UniProt: a hub for protein information The UniProt Consortium}.
\newblock \emph{Nucleic Acids Research}, 2015.
\newblock \doi{10.1093/nar/gku989}.

\bibitem[{Van Der Maaten} and Hinton(2008)]{VanDerMaaten2008}
Laurens {Van Der Maaten} and Geoffrey Hinton.
\newblock {Visualizing data using t-SNE}.
\newblock \emph{Journal of Machine Learning Research}, 2008.
\newblock ISSN 15324435.

\bibitem[Vani and Kumar(2016)]{Vani2016}
K.~Suvarna Vani and K.~Praveen Kumar.
\newblock {Protein fold identification using machine learning methods on
  contact maps}.
\newblock In \emph{CIBCB 2016 - Annual IEEE International Conference on
  Computational Intelligence in Bioinformatics and Computational Biology},
  2016.
\newblock ISBN 9781467394727.
\newblock \doi{10.1109/CIBCB.2016.7758096}.

\bibitem[Whisstock and Lesk(2003)]{Whisstock2003}
James~C. Whisstock and Arthur~M. Lesk.
\newblock {Prediction of protein function from protein sequence and structure},
  2003.
\newblock ISSN 00335835.

\bibitem[Xia et~al.(2017)Xia, Peng, Qi, Mu, Yang, and Tramontano]{Xia2017}
Jiaqi Xia, Zhenling Peng, Dawei Qi, Hongbo Mu, Jianyi Yang, and Anna
  Tramontano.
\newblock {An ensemble approach to protein fold classification by integration
  of template-based assignment and support vector machine classifier}.
\newblock \emph{Bioinformatics}, 2017.
\newblock ISSN 14602059.
\newblock \doi{10.1093/bioinformatics/btw768}.

\bibitem[Zacharaki(2017)]{Zacharaki2017}
Evangelia~I. Zacharaki.
\newblock {Prediction of protein function using a deep convolutional neural
  network ensemble}.
\newblock \emph{PeerJ Computer Science}, 2017.
\newblock ISSN 23765992.
\newblock \doi{10.7717/peerj-cs.124}.

\bibitem[Zemla(2003)]{Zemla2003}
Adam Zemla.
\newblock {LGA: A method for finding 3D similarities in protein structures}.
\newblock \emph{Nucleic Acids Research}, 2003.
\newblock ISSN 03051048.
\newblock \doi{10.1093/nar/gkg571}.

\bibitem[Zhao et~al.(2017)Zhao, Xie, Xu, and Sun]{Zhao2017}
Jing Zhao, Xijiong Xie, Xin Xu, and Shiliang Sun.
\newblock {Multi-view learning overview: Recent progress and new challenges}.
\newblock \emph{Information Fusion}, 2017.
\newblock ISSN 15662535.
\newblock \doi{10.1016/j.inffus.2017.02.007}.

\end{thebibliography}

\onecolumn
\section*{Appendix}

\subsection{Figures}\label{appfigs}
\vspace{2.5mm}

\begin{figure}[htbp]
\setcounter{figure}{0}
\renewcommand{\figurename}{Fig. A} 
\centerline{\includegraphics[width=120mm,height=80mm,keepaspectratio]{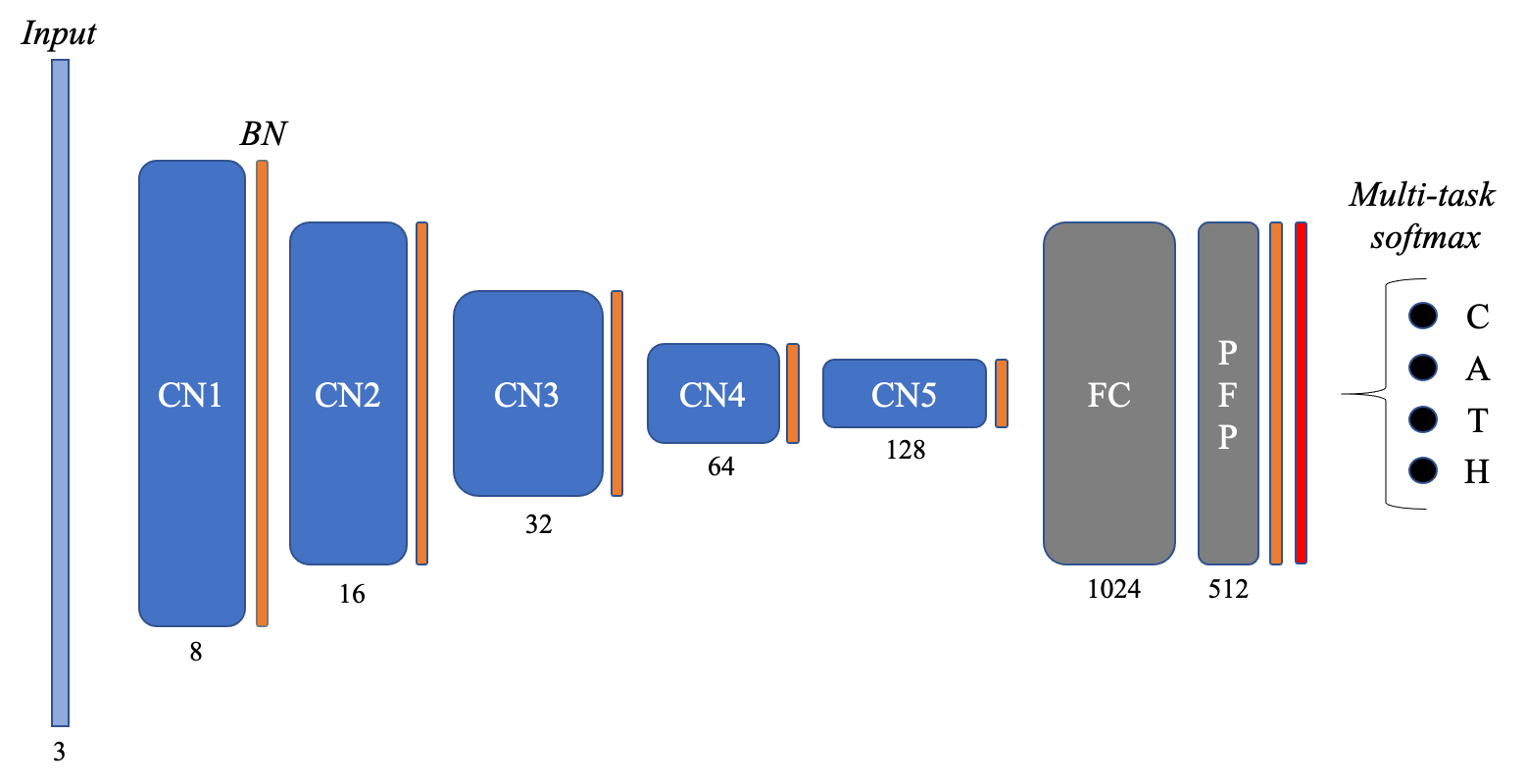}}
\caption{CNN Architecture. A simple CNN was prepared in Keras as a comparator for DenseNet121 as pictured above. Convolutional layers were initialised with glorot uniform distribution, and training was carried out on HRBB as specified for DenseNet121 in \textbf{Section \ref{training}}. CN: Convolutional layer; BN: Batch normalisation; FC: Fully connected (Dense) layer; PFP: Protein fingerprint (Dense) layer; Drop: Dropout layer. CN specifications (kernel, stride): CN1 (4,1); CN2 (4,2); CN3\&4 (4,4); CN5 (4,2).}
\label{figa1}
\end{figure}

\begin{figure*}[htbp]
\renewcommand{\figurename}{Fig. A} 
\centerline{\includegraphics{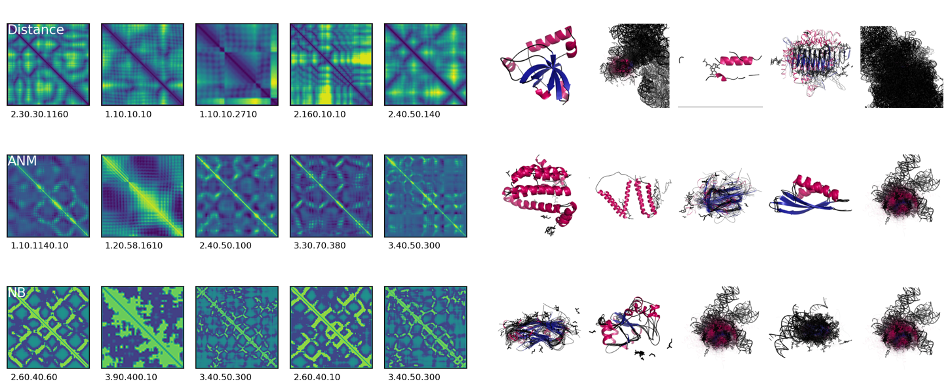}}
\caption{Strongest activation maps (\textit{left}) and corresponding superfamily architectures (\textit{right}) for distance, ANM and NB layers. Top five activations maps (maximum summed intensity) were extracted for the LRCA dataset from the first layer of \textit{DN\_HRBB}. CATH classifications are shown beneath each activation in the format C.A.T.H. Each channel of the first layer responds maximally to different features and domain families. One exception is superfamily 3.40.50.300 (P-loop containing nucleoside triphosphatases), which appears twice for the NB layer and once for the ANM layer, but not for the distance layer.}

\label{fig4}
\end{figure*}

\newpage
\subsection{Tables}\label{Atab}
\vspace{2.5mm}

\begin{table*}[htbp]
\setcounter{table}{0}
\renewcommand{\tablename}{Table A} 
\caption{Prior art in protein structural classification}
\begin{center}

\begin{tabular}{|c|c|c|c|c|}
\hline

\textbf{Model} & \textbf{Representation}& \textbf{N}& \textbf{Task}& \textbf{Performance} \\
\hline
\rowcolor{Gray}\multicolumn{5}{|c|}{\textit{Traditional machine learning approaches}}\\
\hline

\multirow{3}{3cm}{\citet{Shi2009}: SVM}& \multirow{3}{3cm}{Secondary structure features mined from C$_{\alpha}$ distance maps }& \multirow{3}{*}{313} &\underline{SCOP} & \underline{Acc :}\\
& &  & Class (4) & 91\%\\
& & & Fold (27) &51\%-75\% \\
\hline

\multirow{8}{3cm}{\citet{Pires2011}: \\KNN / Random Forest}& \multirow{8}{3cm}{Cut-off Scanning Matrix + SVD from C$_{\alpha}$ distance maps}& &\underline{EC} & \underline{P :}\\
& & 566 & Enzyme superfamily (6) & 99\%\\
& & 55,475&Enzyme subfamily (7) &95\% \\
& & &\underline{SCOP$^{*}$} & \\
& & 110,799& Class& 95\%\\
& & 108,332& Fold& 92\%\\
& & 106,657& Superfamily&93\% \\
& & 102,100& Family&  94\%\\

\hline
\multirow{4}{3cm}{\citet{Taewijit2010}: SVM} & \multirow{4}{3cm}{HMM sequence embeddings + SCCP mined from C$_{\alpha}$ contact maps} & \multirow{4}{*}{2,640} & \multirow{4}{*}{Enzyme subfamily (16)} & \multirow{4}{*}{\underline{Acc :} 73\%-79\%}\\
& &  &  & \\
& & &  & \\
& & &  & \\
\hline

\multirow{4}{3cm}{\citet{Vani2016}: C4.5 Decision Tree + SMOTE} & \multirow{4}{3cm}{Secondary structure features mined from C$_{\alpha}$ distance maps} & \multirow{4}{*}{330} & \multirow{4}{*}{SCOP fold (27)} & \multirow{4}{*}{\underline{F1 :} 72\%}\\
& &  &  & \\
& & &  & \\
& & &  & \\
\hline

\rowcolor{Gray}\multicolumn{5}{|c|}{\textit{Deep CNNs}}\\
\hline
\multirow{5}{3cm}{\citet{Sikosek2019}: \\Pre-trained DenseNet121}& \multirow{5}{3cm}{Heavy atom distances +  NB + ANM}&\multirow{5}{*}{20,798} &\underline{CATH} & \underline{Acc :}\\ 
& & & C (4)& 99\% \\
& & & A (40)&95\% \\
& & & T (1364)&92\% \\
& & & H (2714)&87\% \\
\hline

\multirow{4}{3cm}{\citet{Eguchi2020}: 6-layer CNN with ‘pixel shuffle’ and deconvolution} & \multirow{4}{3cm}{C$_{\alpha}$ distance maps} & \multirow{4}{*}{126,069} & \multirow{4}{*}{CATH: A (40)} & \multirow{4}{*}{\underline{Acc :} 88\%}\\
& &  &  & \\
& & &  & \\
& & &  & \\
\hline

\rowcolor{Gray}\multicolumn{5}{|c|}{\textit{Ensembles}}\\
\hline

\multirow{4}{3cm}{\citet{Zacharaki2017}: Deep CNN ensemble + SVM/KNN} & \multirow{4}{3cm}{Amino acid torsion angles + C$_{\alpha}$ distance maps} & \multirow{4}{*}{44,661} & \multirow{4}{*}{Enzyme superfamily (6)} & \multirow{4}{*}{\underline{Acc :} 90\%}\\
& &  &  & \\
& & &  & \\
& & &  & \\
\hline

\multirow{5}{3cm}{\citet{Newaz2020}: \\Logistic regression}& \multirow{5}{3cm}{Protein structure networks (heavy atoms $\leq$6\si{\angstrom})
+ sequence + GIT (“Concatenate”)
}&\multirow{5}{*}{9,440} &\underline{CATH} & \underline{Acc :}\\ 
& & & C (3)&94\%  \\
& & & A (10)& 87\%-90\%\\
& & & T (14)&88\%-99\% \\
& & & H (5)&93\%-100\% \\
\hline

\multirow{5}{3cm}{This study: \\DenseNet121 ensemble}& \multirow{5}{3cm}{Backbone atom distances +  NB + ANM}&\multirow{5}{*}{15,116-17,048} &\underline{CATH} & \underline{Acc :}\\ 
& & & C (4)&96\%  \\
& & & A (37)& 93\%\\
& & & T (1276)&92\% \\
& & & H (5150)&89\% \\
\hline

\rowcolor{Gray}\multicolumn{5}{|c|}{\textit{Prediction from sequence}}\\
\hline
\multirow{3}{3cm}{\citet{Xia2017}: \\SVM \\+ HMM Ensemble}& \multirow{3}{3cm}{Amino acid sequence}&\multirow{3}{*}{6,451} &\underline{SCOP} & \multirow{3}{*}{\underline{Acc :} 91\%}\\ 
& & & Fold (184)&  \\
& & & & \\

\hline
\multirow{3}{3cm}{\citet{Hou2018}:\\Deep 1D CNN}& \multirow{3}{3cm}{Amino acid sequence}&\multirow{3}{*}{15,956} &\underline{SCOP} & \multirow{3}{*}{\underline{Acc :} 75\%}\\ 
& & & Fold (1,195)&  \\
& & & & \\
\hline

\multicolumn{5}{|c|}{\underline{Abbreviations}. Methods:  ANM: Anisotropic Network Model; GIT: tuned Gauss Intervals; HMM: Hidden Markov Model} \\
\multicolumn{5}{|c|}{KNN: K-Nearest Neighbour; NB: Non-bonded energy; SCCP: Sub-Structural Contact Pattern; SMOTE: Synthetic Majority} \\
\multicolumn{5}{|c|}{Oversampling Technique SVD, Single Value Decomposition; SVM, Support Vector Machine;
N: total size of dataset. Datasets: CATH: } \\
\multicolumn{5}{|c|}{Class, Architecture, Topology, Homolgous superfamily; EC: Enzyme Classification; SCOP: structural classification of proteins. Metrics: } \\
\multicolumn{5}{|c|}{Acc: Accuracy; F1: F1-score; P: precision.$^{*}$Number of categories per class not stated, reference database contains 6, 7, 8 and 24 }\\ 
\multicolumn{5}{|c|}{categories for the four levels of the SCOP hierarchy.} \\
\hline
\end{tabular}
\label{taba1}
\end{center}
\end{table*}

\begin{table*}[htbp]
\renewcommand{\tablename}{Table A} 
\caption{Summary statistics for high-resolution (HR), low-resolution (LR) and NMR datasets}

\begin{center}
\begin{tabular}{|c|c c c c|c c c c c|c|c|c|c|}

\hline
\multirow{2}{*}{\textbf{Atom selection}} & \multicolumn{4}{c|}{\textbf{Instances}} & \multicolumn{5}{c|}{\textbf{Classes}} & \textbf{Res.}$^{*}$ & \textbf{Length}$^{**}$ & \textbf{Length}$^{**}$ & \textbf{Size}$^{**}$ \\
& N$_{total}$ & N$_{train}$ & N$_{val}$ & N$_{test}$ & NC$_{C}$ & NC$_{A}$ & NC$_{T}$ & NC$_{H}$ & NC$_{H<10}$ & (\si{\angstrom}) & (atoms) & (residues) & (Kb)\\
\hline
\rowcolor{Gray}\multicolumn{14}{|c|}{\textbf{\textit{High-Resolution ($\leq$3\si{\angstrom})}}}\\\hline
HRCA & 28,188 & 15,222 & 10,148 & 2,819 & 4 & 41 & 1,276 & 5,129 & 91\% & \multirow{2}{*}{2 (1,3)} & 159 & 161  & 35  \\
HRBB & 28,412&15,342 & 10,228& 2,841& 4& 41& 1,276& 5,150& 91\%& &635 & 161 &167 \\
HRHEAVY & 25,192 & 13,604 & 9,069 & 2,519 & 4 & 41 & 1,269 &5,002 &92\% & & 1234 &106 & 765 \\\hline
\rowcolor{Gray}\multicolumn{14}{|c|}{\textbf{\textit{Low-Resolution ($>$3\si{\angstrom})}}}\\\hline
LRCA & 1,663& \multicolumn{2}{c}{}&1,663 & 4 & 28 & 375 & 885 & 98\% & \multirow{3}{*}{3 (3,4) }&151 & 153  & 176  \\
LRBB & 1,585& \multicolumn{2}{c}{N/A} &1,585 & 4& 28& 369& 859& 98\%& &603  & 153 & 176 \\
LRHEAVY & 1,633 & \multicolumn{2}{c}{} & 1,633 & 4 & 28 & 370 & 873 & 98\%& & 1,173  & 153 & 817 \\\hline

\rowcolor{Gray}\multicolumn{14}{|c|}{\textbf{\textit{NMR}}}\\\hline
NMRCA & 2,902& \multicolumn{2}{c}{}&2,902 & 4 & 26 & 397 & 1,045 & 96\% & \multirow{3}{*}{999 (4,999) }& 94  & 92 & 6 \\
NMRBB & 2,872& \multicolumn{2}{c}{N/A} &2,872 & 4& 26& 396& 1,039& 96\%& & 375 & 92  & 52 \\
NMRHEAVY & 2,875 & \multicolumn{2}{c}{} & 2,875 & 4 & 26 & 395 & 1,036 & 96\%& & 735  & 92 & 238 \\\hline

\multicolumn{14}{|c|}{\underline{Abbreviations}: CA: alpha carbon; BB: backbone; N: Number of instances; NC: Number of classes; NC$_{H<10}$: Proportion of H classes having fewer than }\\ 
\multicolumn{14}{|c|}{ten instances; Res.: Resolution.$^{*}$Mean over all instances of the dataset, (min,max), $^{**}$ Mean  length before pre-processing}\\\hline

\end{tabular}
\label{taba8}
\end{center}
\end{table*}

\begin{table*}[htbp]
\renewcommand{\tablename}{Table A} 
\caption{Impact of atom selection on model performance}
\begin{center}
\begin{tabular}{|c|c c|c c c c|}

\hline
\multirow{2}{*}{\textbf{Representation}} &\multirow{2}{*}{\textbf{N$_{train}$}} &\multirow{2}{*}{\textbf{N$_{test}$}} & \multicolumn{4}{c|}{\textbf{Test Accuracy}} \\
 & & & C & A & T & H \\
\hline
HRCA & 16,913 & 2,360 &94 $\pm$ 0.4\%&82\% $\pm$ 0.6\%&75\% $\pm$ 2.6\%&63\% $\pm$ 2.2\%\\
HRBB & 17,048 & 2,818 &96 $\pm$ 1.5\%&86\% $\pm$ 2.4\%&79\% $\pm$ 1.3\%&67\% $\pm$ 1.7\%\\
HRBB\_DIST\_ONLY & 17,396 & 2,899 &96 $\pm$ 1.2\%&87\% $\pm$ 4.2\%&80\% $\pm$ 2.0\%&69\% $\pm$ 2.4\%\\
HRHEAVY & 15,116 & 2,519 &96 $\pm$ 1.1\%&85\% $\pm$ 0.6\%&77\% $\pm$ 0.5\%&61\% $\pm$ 1.0\%\\
CNN\_HRBB & 17,048 & 2,818 &93 $\pm$ 1.0\%& 79\% $\pm$ 0.8\%&70\% $\pm$ 0.2\%&60\% $\pm$ 0.9\%\\
Benchmark \citep{Sikosek2019} & 12,479 & 8,319 &99\% & 95\%& 92\%& 87\%\\
\hline
\multicolumn{7}{c}{} \\
\hline

\multirow{2}{*}{\textbf{Representation}} &\multirow{2}{*}{\textbf{N$_{train}$}} &\multirow{2}{*}{\textbf{N$_{test}$}} & \multicolumn{4}{c|}{\textbf{F1-score}} \\
 & & & C & A & T & H \\
\hline
HRCA & 16,913 & 2,360 &94 $\pm$ 0.5\%&82\% $\pm$ 0.5\%&73\% $\pm$ 3.1\%&59\% $\pm$ 2.3\%\\
HRBB & 17,048 & 2,818 &96 $\pm$ 1.6\%&86\% $\pm$ 2.6\%&77\% $\pm$ 1.4\%&64\% $\pm$ 2.0\%\\
HRBB\_DIST\_ONLY & 17,396 & 2,899 &96 $\pm$ 1.3\%&87\% $\pm$ 4.5\%&80\% $\pm$ 2.9\%&68\% $\pm$ 4.6\%\\
HRHEAVY & 15,116 & 2,519 &95 $\pm$ 1.2\%&85\% $\pm$ 0.5\%&75\% $\pm$ 0.5\%&59\% $\pm$ 1.0\%\\
CNN\_HRBB& 17,048 & 2,818 &93 $\pm$ 0.9\%&79\% $\pm$ 0.8\%&68\% $\pm$ 0.1\%&57\% $\pm$ 1.7\%\\
\hline
\end{tabular}
\label{taba2}
\end{center}
\end{table*}

\begin{table*}[htbp]
\renewcommand{\tablename}{Table A} 
\caption{HRBB performance across test sets}
\begin{center}
\begin{tabular}{|c|c|c c c c|}

\hline
\multirow{2}{*}{\textbf{Representation}} &\multirow{2}{*}{\textbf{N$_{test}$}} & \multicolumn{4}{c|}{\textbf{Test Accuracy}} \\
 & & C & A & T & H \\
\hline
\rowcolor{Gray}\multicolumn{6}{|c|}{\textbf{\textit{High-Resolution ($\leq$3\si{\angstrom})}}}\\
\hline
HRCA & 2,360 &98\% &92\% &89\% &84\% \\
HRBB  & 2,818 &96\% &86\% &79\% &81\% \\
HRHEAVY  & 2,519 &56\% &26\% &16\% &1\% \\
\hline
\rowcolor{Gray}\multicolumn{6}{|c|}{\textbf{\textit{Low-Resolution ($>$3\si{\angstrom})}}}\\
\hline
LRCA & 1,663 &93\% &80\% &67\% &51\% \\
LRBB  & 1,585 &92\% &79\% &64\% &48\% \\
LRHEAVY  & 1,634 &38\% &13\% &9\% &1\% \\
\hline
\rowcolor{Gray}\multicolumn{6}{|c|}{\textbf{\textit{NMR}}}\\
\hline
NMRCA & 3,047 &91\% &79\% &63\% &46\% \\
NMRBB  & 3,017 &91\% &79\% &61\% &44\% \\
NMRHEAVY  & 3,019 &38\% &7\% &3\% &1\% \\
\hline
Benchmark \citep{Sikosek2019} & 8,319 &99\% & 95\%& 92\%& 87\%\\
\hline
\multicolumn{6}{c}{} \\
\hline
\multirow{2}{*}{\textbf{Representation}} &\multirow{2}{*}{\textbf{N$_{test}$}} & \multicolumn{4}{c|}{\textbf{F1-score}} \\
 & & C & A & T & H \\
\hline
\rowcolor{Gray}\multicolumn{6}{|c|}{\textbf{\textit{High-Resolution ($\leq$3\si{\angstrom})}}}\\
\hline
HRCA & 2,360 &98\% &92\% &88\% &84\% \\
HRBB  & 2,818 &96\% &85\% &77\% &65\% \\
HRHEAVY  & 2,519 &40\% &11\% &4\% &0\% \\
\hline
\rowcolor{Gray}\multicolumn{6}{|c|}{\textbf{\textit{Low-Resolution ($>$3\si{\angstrom})}}}\\
\hline
LRCA & 1,663 &93\% &80\% &67\% &51\% \\
LRBB  & 1,585 &92\% &79\% &64\% &46\% \\
LRHEAVY  & 1,634 &21\% &3\% &2\% &0\% \\
\hline
\rowcolor{Gray}\multicolumn{6}{|c|}{\textbf{\textit{NMR}}}\\
\hline
NMRCA & 3,047 &91\% &79\% &63\% &46\% \\
NMRBB  & 3,017 &90\% &78\% &61\% &44\% \\
NMRHEAVY  & 3,019 &21\% &1\% &0\% &0\% \\
\hline

\multicolumn{6}{c}{} \\
\hline
\multirow{2}{*}{\textbf{Representation}} &\multirow{2}{*}{\textbf{N$_{test}$}} & \multicolumn{4}{c|}{\textbf{PFP homogeneity}} \\
 & & C & A & T & H \\
\hline
\rowcolor{Gray}\multicolumn{6}{|c|}{\textbf{\textit{High-Resolution ($\leq$3\si{\angstrom})}}}\\
\hline
HRCA & 2,360 &84\% &68\% &90\% &94\% \\
HRBB  & 2,818 &83\% &70\% &87\% &92\% \\
HRHEAVY  & 2,519 &0\% &0\% &0\% &0\% \\
\hline
\rowcolor{Gray}\multicolumn{6}{|c|}{\textbf{\textit{Low-Resolution ($>$3\si{\angstrom})}}}\\
\hline
LRCA & 1,663 &55\% &57\% &85\% &93\% \\
LRBB  & 1,585 &44\% &58\% &85\% &94\% \\
LRHEAVY  & 1,634 &0\% &0\% &0\% &0\% \\
\hline
\rowcolor{Gray}\multicolumn{6}{|c|}{\textbf{\textit{NMR}}}\\
\hline
NMRCA & 3,047 &50\% &61\% &82\% &90\% \\
NMRBB  & 3,017 &52\% &60\% &82\% &90\% \\
NMRHEAVY  & 1,634 &0\% &0\% &0\% &0\% \\
\hline
Benchmark \citep{Sikosek2019} & 8,319 &93\% & 89\%& 95\%& 97\%\\
\hline
\end{tabular}
\label{taba3}
\end{center}
\end{table*}

\begin{table*}[htbp]
\renewcommand{\tablename}{Table A} 
\caption{Test set performance of ensemble \textit{DN\_E1}}
\begin{center}
\begin{tabular}{|c|c|c c c c|c c c c|}
\hline
\multirow{2}{*}{\textbf{Atom selection}} & \multirow{2}{*}{\textbf{N$_{test}$}} & \multicolumn{4}{c|}{\textbf{Accuracy}}& \multicolumn{4}{c|}{\textbf{F1-score}}\\
 & &  C & A & T & H & C & A & T & H \\
\hline
\rowcolor{Gray}\multicolumn{10}{|c|}{\textbf{\textit{High-Resolution ($\leq$3\si{\angstrom})}}}\\
\hline
HRCA & 2,360& 96\%& 92\%& 90\%& 84\%& 96\%& 92\%& 89\%& 82\%\\
HRBB & 2,818& 94\%& 90\%& 86\%& 80\%& 94\%&90\% & 86\%& 77\%\\
HRHEAVY &2,519 &94\% & 76\%& 63\%& 58\%&93\% &76\% &61\% &57\% \\
\hline
\rowcolor{Gray}\multicolumn{10}{|c|}{\textbf{\textit{Low-Resolution ($>$3\si{\angstrom})}}}\\
\hline
LRCA & 1,663&90\% &80\% &69\% &53\% &90\% &81\% & 69\%& 51\%\\
LRBB & 1,585&87\% & 78\%& 66\%& 49\%& 87\%& 79\%& 66\%& 47\%\\
LRHEAVY & 1,634& 89\%& 64\%& 50\%& 43\%& 89\%& 66\%& 50\%& 44\%\\
\hline
\rowcolor{Gray}\multicolumn{10}{|c|}{\textbf{\textit{NMR}}}\\
\hline
NMRCA & 3,047 & 88\%& 80\%&65\% &47\% &87\% &81\% &65\% & 47\%\\
NMRBB & 3,017& 86\%& 79\%& 62\%&44\% & 85\%& 79\%& 62\%&43\% \\
NMRHEAVY & 3,019&88\% &64\% &46\% &36\% &87\% &69\% & 51\%& 39\%\\
\hline
BENCHMARK \citep{Sikosek2019} & 12,479 & 99\% & 95\% & 92 \% & 87\% & \multicolumn{4}{c|}{\cellcolor{gray}} \\
\hline
\end{tabular}
\label{taba4}
\end{center}
\end{table*}

\begin{table*}[htbp]
\renewcommand{\tablename}{Table A} 
\caption{Test set performance of ensemble E2}
\begin{center}
\begin{tabular}{|c|c|c c c c|c c c c|}
\hline
\multirow{2}{*}{\textbf{Atom selection}} & \multirow{2}{*}{\textbf{N$_{test}$}} & \multicolumn{4}{c|}{\textbf{Accuracy}}& \multicolumn{4}{c|}{\textbf{F1-score}}\\
 & &  C & A & T & H & C & A & T & H \\
\hline
\rowcolor{Gray}\multicolumn{10}{|c|}{\textbf{\textit{High-Resolution ($\leq$3\si{\angstrom})}}}\\
\hline
HRCA & 2,360& 96\%& 92\%& 90\%& 84\%& 96\%& 92\%& 89\%& 82\%\\
HRBB & 2,818& 96\%& 93\%& 92\%& 89\%& 96\%&94\% & 92\%& 87\%\\
HRHEAVY &2,519 &94\% & 76\%& 63\%& 58\%&93\% &76\% &61\% &57\% \\
\hline
\rowcolor{Gray}\multicolumn{10}{|c|}{\textbf{\textit{Low-Resolution ($>$3\si{\angstrom})}}}\\
\hline
LRCA & 1,663&81\% &54\% &44\% &39\% &81\% &57\% & 52\%& 42\%\\
LRBB & 1,585&79\% & 52\%& 42\%& 37\%& 79\%& 54\%& 51\%& 39\%\\
LRHEAVY & 1,634& 29\%& 3\%& 40\%& 40\%& 13\%& 4\%& 47\%& 42\%\\
\hline
\rowcolor{Gray}\multicolumn{10}{|c|}{\textbf{\textit{NMR}}}\\
\hline
NMRCA & 3,047 & 78\%& 56\%&39\% &33\% &78\% &57\% &47\% & 37\%\\
NMRBB & 3,017& 83\%& 57\%& 41\%&34\% & 82\%& 56\%& 48\%&37\% \\
NMRHEAVY & 3,019&28\% &6\% &38\% &32\% &12\% &5\% & 45\%& 36\%\\
\hline
BENCHMARK \citep{Sikosek2019} & 12,479 & 99\% & 95\% & 92 \% & 87\% & \multicolumn{4}{c|}{\cellcolor{gray}} \\
\hline
\end{tabular}
\label{taba5}
\end{center}
\end{table*}

\begin{table*}[t]
\renewcommand{\tablename}{Table A} 
\caption{Per category performance of HRBB model on HRBB test set}
\begin{center}
\begin{tabular}{|c c c c c c|}
\hline
\textbf{CATH label} & \textbf{Description} & \textbf{Precision} & \textbf{Recall} & \textbf{F1-score} & \textbf{Support}\\
\hline
\rowcolor{Gray}\multicolumn{6}{|c|}{Class}\\
\hline
1 & Mainly alpha & 97\%& 96\%& 97\%& 655\\
2 & Mainly beta & 94\%& 94\%& 94\%& 589\\
3 & Alpha - beta & 97\%& 97\%& 97\%& 1554\\
4 & Few secondary structures & 50\%& 55\%& 52\%& 20\\
\hline
\rowcolor{Gray}\multicolumn{6}{|c|}{Architecture}\\
\hline
1.10 & Orthogonal bundle& 86\%&90 &88 &389 \\
1.20 & Up-down bundle & 82\%& 74\%& 78\%& 206\\
1.25 & Alpha horseshoe & 88\%& 88\%& 88\%& 49\\
1.40 & Alpha solenoid & 100\%& 100\%& 100\%&1 \\
1.50 & Alpha / alpha barrel & 91\%& 100\%& 95\%& 10\\
\hline
2.10 & Ribbon & 64\%& 80\%& 71\%& 20\\
2.20 & Single sheet & 47\%& 38\%& 42\%& 24\\
2.30 & Roll &83\% &75\% &79\% &73 \\
2.40 & Beta barrel &84\% &86\% &85\% &137\\
2.50 & Clam &100\% &50\% &67\% &2 \\
2.60  & Sandwich & 98\%& 94\%& 96\%& 250\\
2.70 & Distorted sandwich& 73\%& 92\%& 81\%& 12\\
2.80 & Trefoil & 92\%& 100\%& 96\%& 12\\
2.90 & Orthogonal prism& 100\%& 100\%& 100\%& 2\\
2.100 & Aligned prism& 100\%& 67\%& 80\%& 3\\
2.102 & 3-layer sandwich & 50\%& 100\%& 67\%& 1\\
2.105 & 3 propeller & 0\%& 0\%& 0\%& 0\\
2.110 & 4 propeller& 0\%& 0\%& 0\%& 0\\
2.115 & 5 propeller& 67\%& 100\%& 80\%& 4\\
2.120 & 6 propeller& 100\%& 83\%& 91\%& 6\\
2.130  & 7 propeller& 100\%& 100\%& 100\%& 14\\
2.140 & 8 propeller& 100\%& 100\%& 100\%& 1\\
2.150 & 2 solenoid & 100\%& 100\%& 100\%& 1\\
2.160 & 3 solenoid & 100\%& 93\%& 96\%& 14\\
2.170 & Beta complex & 47\%& 62\%& 53\%& 13\\
2.180 & Shell & 0\%& 0\%& 0\%& 0\\
\hline
3.10 & Roll & 76\%& 70\%& 73\%& 105\\
3.15 & Super roll & 100\%& 100\%& 100\%& 3\\
3.20 & Alpha-beta barrel & 93\%& 51\%& 66\%& 128\\
3.30 & 2-layer sandwich& 83\%& 86\%& 84\%& 399\\
3.40 & 3-layer (aba) sandwich & 89\%& 97\%& 93\%& 702\\
3.50 & 3-layer (bba) sandwich & 100\%& 89\%& 94\%& 28\\
3.55 & 3-layer (bab) sandwich& 100\%& 100\%& 100\%& 2\\
3.60 & 4-layer sandwich & 96\%& 72\%& 82\%& 32\\
3.65 & Alpha-beta prism & 100\%& 100\%&100\% &2 \\
3.70 & Box & 100\%& 100\%& 100\%& 3\\
3.75 & 5-stranded propeller & 100\%& 100\%& 100\%& 2\\
3.80 & Alpha-beta horseshoe & 100\%& 91\%& 95\%& 11\\
3.90 & Alpha-beta complex & 77\%& 79\%& 78\%& 137\\
3.100 & Ribosomal protein L15& 0\%& 0\%& 0\%&0 \\
\hline
4.10 & Irregular & 41\%& 55\%& 47\%& 20\%\\
\hline
\end{tabular}
\label{taba7}
\end{center}
\end{table*}

\begin{table*}[t]
\renewcommand{\tablename}{Table A} 
\caption{Applying a Random Forest ensemble to protein fingerprints}
\begin{center}
\begin{tabular}{|c c|c|c c c c|}
\hline
\multirow{2}{*}{\textbf{Training set}} & \multirow{2}{*}{\textbf{Test set}}& \multirow{2}{*}{\textbf{N$_{test}$}}& \multicolumn{4}{c|}{\textbf{Accuracy}} \\
 &  &  &C & A & T &H \\
\hline
HRCA & HRCA & 2,360 & 95\% & 82\%& 66\%& 47\%\\
HRBB\_DIST\_ONLY &HRBB\_DIST\_ONLY & 2,899& 97\%& 87\%& 70\%& 52\%\\
HR\_HEAVY & HRHEAVY & 2,519& 95\%& 82\%& 65\%& 43\%\\
\hline
\multirow{9}{*}{HRBB} & HRCA & 2,360 & 98\% & 88\%& 72\%& 52\%\\
 & HRBB & 2,818 &96\% &85\% & 68\%& 48\% \\
 & HRHEAVY & 2,519 &56\% & 26\%& 16\%& 2\% \\
 & LRCA & 1,663 & 93\%&79\%& 61\%& 41\%\\
 & LRBB & 1,585 & 92\%& 78\%& 61\% &40\%\\
 & LRHEAVY & 1,634 & 38\%&17\% & 10\%&6\%  \\
 & NMRCA & 3,047 & 91\%& 81\%&66\% &51\%\\
 & NMRBB & 3,017 & 92\%& 80\%& 65\%& 49\%\\
 & NMRHEAVY & 3,019 & 38\%& 23\%& 7\%& 4\% \\
\hline
HRBB (DenseNet121) & HRBB& 2,360& 96\%& 92\%& 90\%& 84\%\\
\hline
\multicolumn{7}{|c|}{Feature vectors were extracted for each test set using a model pre-trained on CA, BB }\\
\multicolumn{7}{|c|}{(3-part representation), BB (distance only), or heavy atom HR training sets. A Random  }\\
\multicolumn{7}{|c|}{Forest model from the scikit-learn ensembles module was then trained and evaluated on }\\
\multicolumn{7}{|c|}{each set using 10-fold cross-validation. Random Forest (n\_est$=$150, max\_depth$=$50) was }\\
\multicolumn{7}{|c|}{selected on the basis of comparison with linear SVC, polynomial SVC and logistic }\\
\multicolumn{7}{|c|}{regressors.}\\

\hline
\end{tabular}
\label{taba6}
\end{center}
\end{table*}

\end{document}